\documentclass[journal]{IEEEtran}

\usepackage{mathrsfs}
\usepackage{epsfig}
\usepackage{epstopdf}

\usepackage{amssymb}
\usepackage{amsthm}
\usepackage{graphics} 
\usepackage{indentfirst}
\usepackage{amsfonts} 

\usepackage{cite}
\usepackage{flushend}
\usepackage[cmex10]{amsmath}
\usepackage{caption}
\usepackage{array}
\usepackage{cases}
\usepackage{mdwmath}
\usepackage{mdwtab}
\usepackage{eqparbox}
\usepackage{fixltx2e}

\usepackage{makecell} 
\usepackage[thicklines]{cancel}
\usepackage{booktabs}
\usepackage{multirow}
\usepackage{textcomp}
\usepackage{threeparttable}
\usepackage{subcaption}

\usepackage{color}

\usepackage{diagbox}

\newtheorem{remark}{\rm{\textbf{Remark}}}

\providecommand{\diag}{{\rm diag}}

\providecommand{\beq}{\begin{equation}}
\providecommand{\eeq}{\end{equation}}
\providecommand{\bea}{\begin{eqnarray}}
\providecommand{\eea}{\end{eqnarray}}

\usepackage{algpseudocode,algpascal}
\usepackage[ruled,linesnumbered]{algorithm2e}
\usepackage{etoolbox}
\makeatletter
\patchcmd{\@algocf@start}
  {-1.5em}
  {0pt}
  {}{}
\makeatother

\begin{document}

\title{Generation of Uncorrelated Residual Variables for Chemical Process Fault Diagnosis via Transfer Learning-based Input-Output Decoupled Networks}

\author{Zhuofu Pan, Qingkai Sui, Yalin Wang,~\IEEEmembership{Senior Member,~IEEE}, Jiang Luo,\\ Jie Chen, and Hongtian Chen,~\IEEEmembership{Member,~IEEE}
\thanks{

This work was supported in part by the National Natural Science
Foundation of China (92267205, 62103063), in part by the Open Project of Xiangjiang Laboratory (22XJ03019),
and in part by the Scientific Research Fund of Hunan Provincial Education Department (22A0459). (\textit{Corresponding author: Qingkai Sui; Yalin Wang.}) }
\thanks{Zhuofu Pan is with the Xiangjiang Laboratory and the School of Microelec-tronics and Physics,
Hunan University of Technology and Business, Changsha 410205, China, and also with the School of Automation,
Central South University, Changsha 410083, China (e-mail: joffpan\_ai@outlook.com).}
\thanks{Qingkai Sui, Yalin Wang, and Jiang Luo are with the School of Automation,
Central South University, Changsha 410083, China
(e-mail: suiqingkai@csu.edu.cn; ylwang@csu.edu.cn; ljiang@csu.edu.cn).}
\thanks{Jie Chen is with the School of Computer Science, Hunan University of Technology and Business, Changsha 410205, China (e-mail: cj1732@126.com).}
\thanks{Hongtian Chen is with the Department of Automation, Shanghai Jiao Tong University, Shanghai 200240, China (e-mail: hongtian.chen@sjtu.edu.cn).}

}

\markboth{ }%
{Shell \MakeLowercase{\textit{et al.}}: A Sample Article Using IEEEtran.cls for IEEE Journals}

\maketitle
\begin{abstract}
    Structural decoupling has played an essential role in model-based fault isolation and estimation in past decades,
    which facilitates accurate fault localization and reconstruction thanks to the diagonal transfer matrix design.
    However, traditional methods exhibit limited effectiveness in modeling high-dimensional nonlinearity and big data, and the decoupling idea has not been well-valued in data-driven frameworks.
    Known for big data and complex feature extraction capabilities, deep learning has recently been used to develop residual generation models. Nevertheless, it lacks decoupling-related diagnostic designs.
    To this end, this paper proposes a transfer learning-based input-output decoupled network (TDN) for diagnostic purposes, which consists of an input-output decoupled network (IDN) and a pre-trained variational autocoder (VAE).
    In IDN, uncorrelated residual variables are generated by diagonalization and parallel computing operations.
    During the transfer learning phase, knowledge of normal status is provided according to VAE's loss and maximum mean discrepancy loss to guide the training of IDN.
    After training, IDN learns the mapping from faulty to normal, thereby serving as the fault detection index and the estimated fault signal simultaneously.
    At last, the effectiveness of the developed TDN is verified by a numerical example and a chemical simulation.

\end{abstract}



\begin{IEEEkeywords}
    Transfer learning, input-output decoupled networks, faulty-to-normal mapping, fault detection and estimation
\end{IEEEkeywords}

\section{Introduction}\label{Sec1}
During the past decades, model-based fault diagnosis technologies have made significant contributions to maintaining the safe, reliable, and efficient operation of chemical processes \cite{chen2021data,9072621}.
By abstracting systems into parametric and perturbed forms, they can effectively model system behavior and monitor intolerable faults.
However, as the production scale expands and industrial data grows exponentially, traditional fault diagnosis methods face difficulties in accurate modeling and efficient data processing \cite{10008204,Ma8648380}.
Data-driven fault diagnosis methods, including shallow learning methods (represented by multivariate statistical analysis and kernel representation) and deep learning methods, have been developed in recent years to alleviate these problems, which have received tremendous attention from industrial communities \cite{9882010}.


Shallow learning-based fault detection (FD) methods, such as canonical correlation analysis \cite{Chen7997784}, principal component analysis (PCA) \cite{zhang2022turning,Jiang7185395}, kernel PCA \cite{liu2022intelligent}, and support vector machines \cite{kubik2022smart} have grown substantially over the last century.
They can detect faults occurring in simple systems well but struggle to handle big data problems for large-scale complex systems, especially in today's era of readily available data \cite{long2022novel}.

Specializing in extracting complex features, deep learning emerged as a milestone in data-driven development.
Since the bottleneck of its training difficulties was broken in 2006 \cite{Hinton2006}.
It has shown success in several fields, including natural language processing, image recognition, and also, fault diagnosis \cite{Pan9768200,TIAN2022108466}.
For example, 
Yu et al. \cite{YU202147} proposed an intensified iterative learning model for FD, where the input to the current hidden layer is derived from all previously hidden layers to avoid information loss.
Tang et al. \cite{TANG2021444} developed a quality-related fault detection method based on the deep variational information bottleneck and variational autoencoder (VAE) to identify whether the fault is related to quality variables. 
Zhang et al. \cite{zhang2020industrial} presented a recurrent Kalman VAE for extracting dynamic and nonlinear features implicit in industrial processes.
Although many deep learning-based FD techniques have been discussed, scholars exhibit less interest in other two fault diagnosis tasks.

In addition to the FD objective for monitoring intolerable system abnormalities, fault isolation (FI), and fault estimation/identification (FE) are also critical subsequent tasks in fault diagnosis.
FI aims to locate the root cause of a system failure, which can be achieved by a model-based decoupled transfer function design or a data-driven contribution assessment scheme (e.g., contribution plots \cite{westerhuis2000generalized}, reconstruction-based contributions \cite{alcala2009reconstruction}, etc.).
Ding believes that perfect FI is the dual form of the decoupling control problem \cite{DingSX2008book}.
Unfortunately, this idea is less represented in data-driven frameworks.
The main purpose of FE is to estimate or reconstruct the fault signal implicit in the observation. In model-based FE schemes, the mainstream is to expand the system parameters or states as the faults to be estimated \cite{tidriri2016bridging}.
Since decoupled residual signals can provide correct fault directions and promote estimation error reduction on non-faulty variables, structural decoupling is always a hot topic in model-based fault estimator designs \cite{liu2022fault}.
Nevertheless, data-driven, especially deep learning-based structural decoupling and FE methods are greatly absent \cite{witczak2017neural}.

Transfer learning (TL) is one of the branches of deep learning developed in recent years.
It is based on the idea that knowledge in similar domains can be shared for learning, leading to fast training and improved accuracy \cite{chen2022transfer}.
Most of TL-based fault diagnosis methods focus on transferring data or knowledge in multi-operation, multi-modal, or multi-block scenarios.
For instance,
Wu et al. \cite{WU2020106731} proposed a transfer learning-based multimode fault detection method, that utilizes source domain data to expand other data-insufficient modalities.
He et al. \cite{he2022deep} introduced a deep multi-signal fusion adversarial model to transfer knowledge between different working conditions of the axial piston pump.
Cheng et al. \cite{cheng2022transfer} proposed a VAE-based federal neural network to detect system fault before and after performance degradation.
Although many TL-based fault diagnosis methods have been successfully developed, fewer studies investigate state migration (from faulty to fault-free) for the FE purpose.

Based on two fundamental issues: 1) deep learning-based residual decoupler and 
2) deep learning-based state migration mapping,
this study proposes a transfer learning-based input-output decoupled network (TDN) for fault diagnosis objectives.
It has the following three-fold contributions.
\begin{enumerate}
\item An input-output decoupled network (IDN) is designed to generate uncorrelated residual variables for FD and FE purposes;
\item A TL-based learning framework is proposed to assist IDN in learning the mapping from faulty to normal via losses from the pre-trained VAE and maximum mean discrepancy;
\item The effectiveness of the decoupling design is verified by visualizing the covariance matrix. TDN yields more accurate FE results than the undecoupled one.
\end{enumerate}

The rest of the paper is organized as follows.
Section \ref{Sec2} introduces the basic principle of VAE and formulates the FD and FE problems. 
Section \ref{Sec3} proposes IDN and TL-based VAE framework to construct TDN.
Then, the TDN residual generator is developed for FD and FE purposes.
Section \ref{Sec4} verifies the effectiveness of the proposed method via a numerical example and a three-tank system (TTS).
Finally, Section \ref{Sec5} concludes this study.

\emph{Notations:} All notations in this study are generic.
$m_x$ denotes the number of variables of $x$;
$\mathbb{R}^{m_x}$ represents the real $m_x$-dimensional space;
$(k)$ denotes the $k^{th}$ collected sample;
$N_x$ represents the total size of $x$;
$\mathbb{D}_x$ denotes the feasible domain of $x$;
operation $\circ$ stands for the function composition;
$\mathcal{B}$, $\mathcal{U}$, and $\mathcal{N}$ stand for Bernoulli, uniform, and normal distribution, respectively.
The superscripts $n$ and $f$ indicate that the variable is in a normal or faulty state.

\section{Preliminaries and Problem Formulation}\label{Sec2}

\subsection{Variational Autoencoder}\label{Sec2:A}
VAEs can be considered a type of probabilistic deep autoencoder.
It encodes the distribution parameters of latent variables instead of compressed deterministic features, and then reconstructs the observation using a decoder.
Its remarkable ability to fit the observation distribution has made it one of the leading generative models.
\begin{figure}[!ht]\centering
	\includegraphics[width=8.3cm]{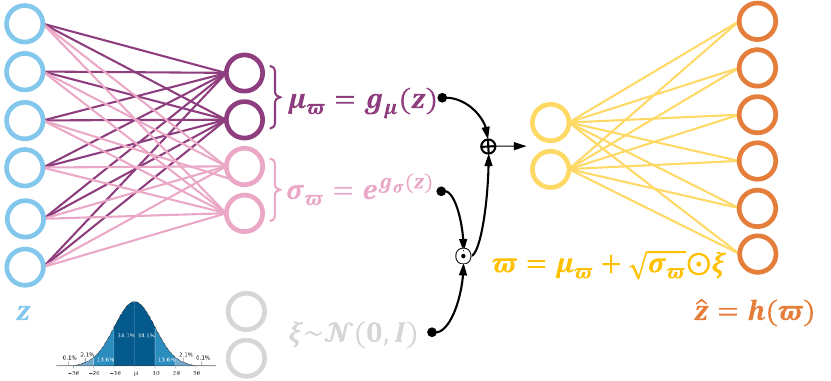}
	\caption{The model structure of VAEs.}\label{fig2-1}
\end{figure}

Fig. \ref{fig2-1} displays the basic structure of VAEs.
The forward propagation of VAEs can be expressed by

\begin{equation}\label{eq2-2}
	\begin{split}
	\rm{encoding:}\  & \mu_\varpi  =g_\mu (z;\theta_\mu ),
    		    \sigma^2_{\varpi} =e^{g_\sigma (z;\theta_\sigma )}\\
    \rm{sampling:}\ &\varpi_s = \mu_\varpi  + {\sigma_\varpi } \odot \xi_s   \\
    \rm{decoding:}\ &\hat{z}= \frac{1}{N_s} \sum_{s=1}^{N_s} h(\varpi_s;\theta_h),
	\end{split}
\end{equation}	
where $z$ and $\varpi $ represent the observation and latent variables, respectively;
$g_\mu(\cdot)$, $g_\sigma (\cdot)$, and $h(\cdot)$ denote three fully connected neural networks (FCNNs),
separately outputting the mean and logarithmic variance of latent variables, and the reconstructed observation;
$\theta_\mu $, $\theta_\sigma  $, and $\theta_h $ are their parameters to be optimized;
$ \xi \sim \mathcal{N} (0,I)$ is a noise introduced to perform reparameterization tricks;
$\varpi_s$ represents the $s^{th}$ sampled latent variables in Monte Carlo sampling.

Assume that an observation $z$ is generated from latent variables $\varpi$ following a prior distribution $p(\varpi) = \mathcal{N}(0,I)$;
its likelihood $p(z|\varpi)$ obeys a Gaussian distribution, satisfying
\begin{equation}\label{eq2-0}
    \hat z_x = h(\varpi _s) + \zeta,\ p(z|\varpi) = \mathcal{N} (h(\varpi), \varrho  I),
\end{equation}
where $\zeta \sim \mathcal{N}(0, \varrho I)$;
$\hat z _s$ represents the $s^th$ reconstructed observation;
$\varrho$ denotes a positive coefficient;
$I$ is a unit matrix with a suitable size.

According to the Bayes' theorem, the posterior $p(\varpi|z)$ can be expressed by
\begin{equation}\label{eq2-1}
    p(\varpi|z) = \frac{p(z,\varpi)}{p(z)} = \frac{p(z|\varpi)p(\varpi)}{\int p(z|\varpi)p(\varpi)\,d\varpi}.
\end{equation}
Note that $p(\varpi|z)$ cannot be represented explicitly since $p(z)$ is an unknown complicated distribution.
Instead, VAEs approximate the posterior using another distribution $q_z(\varpi) =\mathcal{N}(g_\mu (z),Ie^{g_\sigma (z)})$, dedicated to reducing the Kullback-Leibler (KL) divergence between them

\begin{equation}\label{eq2-1-0}
    KL(q_z(\varpi ),p(\varpi |z)) = \log p(z) - \mathbb{E}_{\varpi \sim q_z}\left[ \log \frac{p(z,\varpi)}{q_z(\varpi )} \right].
\end{equation}
The objective in (\ref{eq2-1-0}) is equivalent to \cite{pan2023vae}
\begin{equation}\label{eq2-1-1}
    \begin{split}
        &\min KL(q_z(\varpi ),p(\varpi |z)) = \max \mathbb{E}_{\varpi \sim q_z}\left[ \log \frac{p(z,\varpi)}{q_z(\varpi )} \right]\\
        &= \max \mathbb{E} {_{\varpi \sim q_z}}\left[ {\log p(z|\varpi )} \right] - \mathbb{E} {_{\varpi \sim q_z}}\left[ {\log \frac{{{q_z}(\varpi )}}{{p(\varpi )}}} \right]\\
        &= \min \mathbb{E} {_{\varpi \sim q_z}} \left[ \| z - \hat z\|^2_2 \right] + 2\varrho KL({q_z}(\varpi )||p(\varpi )).
    \end{split}
\end{equation}
Thus, we can deduce the loss function of VAEs $\mathcal{J}_\mathcal{V}$ \cite{doersch2016tutorial}
\begin{equation}\label{eq2-1-2}
    \begin{split}
        &{\mathcal{J}_\mathcal{V} } = \frac{1}{N_k}\frac{1}{N_s}\sum\limits_{k = 1}^{N_k} {  \sum\limits_{s=1}^{N_s} \left\| {z(k) - \hat z_s(k)} \right\|_2^2 + \lambda_\mathcal{V} {\mathcal{J} _{kl}}(k) } \\
        &{\mathcal{J} _{kl}}(k) = \frac{1}{2}\sum\limits_{j = 1}^{{m_\varpi }} {\left( {\mu _j^2(k) + \sigma_j^2(k) - 1 - \log \sigma_j^2(k)} \right)},
    \end{split}
\end{equation}
where $\lambda_\mathcal{V} $ is a positive constant.
It can be observed that, in addition to minimizing reconstruction error, VAEs also impose constraints on the posterior distribution of latent variables, which demonstrates a belief trade-off between the data and prior.
\subsection{Problem Formulation}
Assume the process variables $x^n$ (in normal status) affected by a fault $f$ can be expressed by \cite{QinSJ2012}
\begin{equation}\label{eq2-6}
    \begin{split}
    x^f&=x^n+f\\
    f &= \varOmega  \mathfrak{f}, 
    \end{split}
\end{equation}
where $x^n$ denotes the normal sample before adding the fault $f$, which is unobservable.
Process fault $f$ can be expressed as the multiplication of the direction $\varOmega$ and magnitude $\Vert \mathfrak{f}\Vert _2$, whose definition can be find in \cite{QinSJ2012}.

The purpose of FD is to develop a behavior simulator to that can learn the operating characteristics of the system in normal conditions, as well as an associated residual generator that can produce residuals $\phi=\varPhi (z)$ for detecting faults.
$\varPhi: \mathbb{D}_{Z^f} \to \mathbb{D}_{{\phi}}$ denotes the mapping from $Z^f$ to the residuals ${\phi}$.
For the fault-free sample $z^n$ and faulty sample $z^f$, the following objectives are expected to achieve 
\begin{equation}\label{eq2-7}
    \begin{split}
        \mathbb{E}[\varPhi(z^n)]=0,\ \Vert\varPhi(z^n) \Vert _2\leq \varepsilon_{\phi},\ \mathbb{E}[\varPhi(z^f)] \ne 0.
    \end{split}
\end{equation}
``$(k)$'' is omitted here and after for notation simplicity. ``$\varepsilon_{\phi}$'' is some small positive number.

If the fault impact on each residual variable is separable, we can represent it in a decoupled form \cite{DingSX2020book}

\begin{equation}\label{eq2-8}
	\phi=\varPhi(z^f) =\begin{bmatrix}
        \varPhi _1(z^f_{1},0,\dots,0)\ \ \\
        \varPhi _2(0,z^f_{2},0,\dots ,0) \\
        \vdots\\
        \varPhi _{m_{\phi}}(0,\dots , 0,z^f_{m_{\phi}})
        \end{bmatrix}
\end{equation}
where ${\phi _j}=\varPhi _j (z^f)$ denotes the sub-mapping of the $j^{th}$ residuals variable.
$m_{\phi}$ stands for the dimension of $z^f$.

Based on the decoupled residuals, it is not difficult to establish the fault estimator $\varXi _{j} (\cdot)$ for $f_{j}$.
It can be defined by
\begin{equation}\label{eq2-9}
    \tilde{f}_j=\varXi _{j}(z^f_j)=z^f_j - \tilde{z}^n_j
\end{equation}
where $\varXi _j: \mathbb{D}_{z^f_j} \to \mathbb{D}_{\tilde{f}_j}$ denotes the fault estimator for fault $f_j$ to be learned \cite{DingSX2020book};
$\tilde{f}_j$ is the estimated fault signal on variable $j$.


\section{Transfer Learning-Based Input-Output Decoupled Network}\label{Sec3}
In order to obtain uncorrelated residual variables, this section proposes IDN.
It ensures that a single output variable is only affected by a corresponding input variable.
Then, a TL-based framework is developed to assist in training of IDN, allowing it to map faulty samples to normal ones and perform FE functionality.

\subsection{Input-Output Decoupled Network}\label{Sec3:A}
To achieve the decoupling objective,
the $j^{th}$ output of the model should only be related to the $j^{th}$ input and not to other variables.
Define $z_j^-$ to be the observation where all but the $j^{th}$ variable are zero, denoting by
\begin{equation}\label{eq3-1}
    \begin{split}
        z &= (z_1,\cdots,z_{j-1},z_j,z_{j+1},\cdots,z_{m_z})\\
        z_j^-&=(0,\cdots,0,z_j,0,\cdots,0)
    \end{split}
\end{equation}

\begin{figure}[!ht]\centering
	\includegraphics[width=7.5cm]{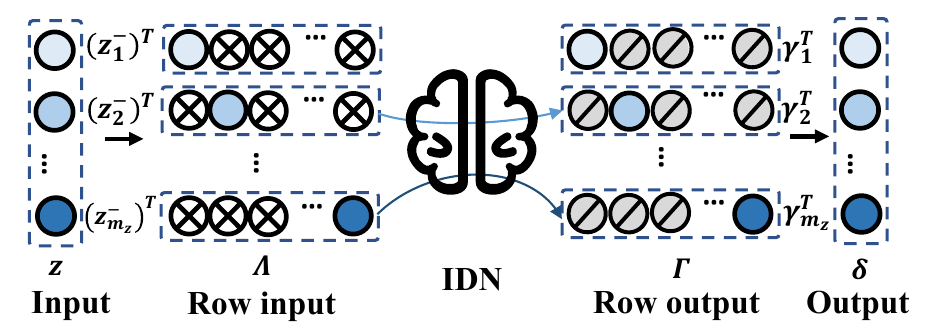}
	\caption{Calculation flow of IDN for an individual sample.
    The notation ``$\otimes$'' marks the zero elements; ``$\oslash$'' stands for the non-adopted elements.}\label{fig3-1}
\end{figure}

Then, the decoupled model $\mathcal{D}(\cdot )$ should satisfy the following condition

\begin{equation}\label{eq3-2}
    \mathcal{D}_j(z)=\mathcal{D}_j(z_j^-),j=1,2,\cdots,m_z
\end{equation}

where $\mathcal{D}_j(z)$ denotes the $j^{th}$ output of the model $\mathcal{D}(\cdot )$.
The primary concern is how to construct a model that can not only provide decoupled input-output but also accurately simulate the system mapping.
\begin{remark}
Note that in neural networks, parallel computation is decoupled among samples.
This means that the network outputs of each sample are not correlated,
but share the same mapping or distribution through  the learned model.
\end{remark}
Inspired by the parallelism property, we designed IDN, which has the structure shown in Fig. \ref{fig3-1}.

In the forward propagation of IDN,
the input variables $z \in \mathbb{R} ^{m_z}$ undergoes diagonalization to transform it into a diagonal matrix $\varLambda  \in \mathbb{R} ^{m_z\times m_z}$, which can be described by
\begin{equation}\label{eq3-4}
    \varLambda=\diag(z) =
    \begin{pmatrix}
    (z_1^-)^T     \\
    (z_2^-)^T     \\
    \vdots \\
    (z_{m_z}^-)^T
    \end{pmatrix},\
    \varLambda_{ji}=
    \begin{cases}
        z_j, & \text{ if } i=j \\
        0, & \text{ if } i\neq j,
    \end{cases}
\end{equation}
where $(z^{-}_j)^T$ defined by \eqref{eq3-1} denotes the $j^{th}$ row in the matrix $\varLambda $;
$\varLambda_{ji}$ denotes the element in row $j$, column $i$.
As a result, each row variable $(z_j^-)^T$ can be treated as a separate sample for parallel forward operations,
and their output vectors are uncorrelated with each other.

\begin{figure}[t]\centering
	\includegraphics[width=6.5cm]{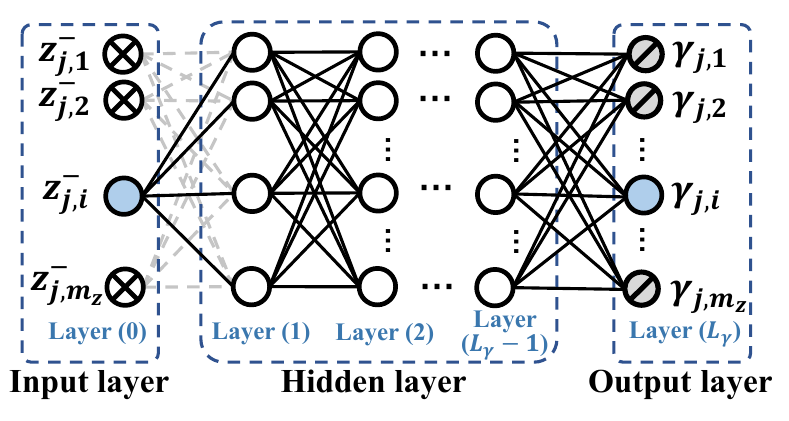}
	\caption{The detailed structure of IDN.}\label{fig3-2}
\end{figure}

The output vectors $\{\gamma ^T_j\}_{j=1}^{m_z}$ of each row inputs can be defined by
\begin{equation}\label{eq3-5}
	\gamma_j = \mathcal{D}_\gamma  (z^- _j; \theta_ \gamma ),
\end{equation}
where $\gamma_j \in \mathbb{R} ^{m_z}$ is only associated with the variable $z_j$;
the nonlinear mapping in decoupling network $\mathcal{D}_\gamma: z^-_j \to \gamma_j$ 
can be a fully connected neural network (FCNN), whose parameters $\theta _{\gamma }$ are shared for $z_j^-,j=1,2,\dots,m_z$;
An FCNN to construct $\mathcal{D}_{\gamma } (\cdot )$ can be expressed by
\begin{equation}\label{eq3-6}
    \begin{split}
        \mathcal{D}_\gamma &= \mathcal{A}^{(L_\gamma)} \circ \mathcal{L}^{(L_\gamma)} \circ \cdots \circ \mathcal{A}^{(1)} \circ \mathcal{L}^{(1)}\\
        \iota ^{(l)} &= \mathcal{L}^{(l)}(a^{(l-1)}) = W^{(l)} a^{(l-1)} + b^{(l)} \\
        a ^{(l)} &= \mathcal{A}^{(l)}(\iota  ^{(l)}),
    \end{split}
\end{equation} 
where $(l) = 1,2,\dots,L_\gamma $ denotes the index of layers;
$L_\gamma $ is the number of layers in FCNN;
$\mathcal{A}$ and $\mathcal{L}$ denotes the nonlinear activation mapping and linear weighting mapping, respectively;
$W^{(l)}\in \mathbb{R}^{m_{(l)} \times m_{(l-1)}}$ represents the weight matrix between layer $(l)$ and $(l-1)$;
$b^{(l)}$ denotes the bias in the $(l)^{th}$ layer;
$\theta_\gamma = \{W^{(l)}, b^{(l)} \}_{l=1}^{L_\gamma }$;
$\iota ^{(l)}$ and $a  ^{(l)}$ denote the weighted input and activation values in layer $(l)$, respectively.
In particular, $a  ^{(0)} = z^-_j \in \mathbb{R}^{m_z}$, $a  ^{(L_\gamma )} = \gamma_j \in \mathbb{R}^{m_z}$, $j = 1,2,\dots,m_z$.
The detailed structure of FCNN can be found in Fig. \ref{fig3-2},
in which $z_{j,i}^-$ and $\gamma_{j,i}$ represents the $i^{th}$ variable in $z_j^-$ and $\gamma_j$, respectively.

In order to maintain identical dimensions of the input $z$ and output $\gamma $ of IDN, a mapping
$\mathcal{D}_\delta = [\mathcal{D}_{\delta,1}\ \mathcal{D}_{\delta,2}\ \dots\ \mathcal{D}_{\delta,m_z} ]^T$ is applied to convert
matrix $ \varGamma   \in \mathbb{R}^{m_z\times m_z}$ into vector $\delta \in \mathbb{R}^{m_z}$, satisfying
\begin{equation}\label{eq3-7}
	\delta _j = \mathcal{D}_{\delta ,j} (\gamma _j).
\end{equation}
Here, $\mathcal{D}_{\delta ,j}(\cdot)$ can be any operation regarding $\gamma  _j$,
e.g., $\mathcal{D}_{\delta ,j}(\gamma _j) \\= \sum _{i=1}^{m_z} \gamma _{j,i}$.
In this study, we use the inverse operation of the diagonal function $\diag^{-1}(\cdot)$ to take the main diagonal of the matrix to obtain
\begin{equation}\label{eq3-8}
	\delta =\mathcal{D}_{\delta}(\varGamma)= \diag^{-1}(\varGamma),\ \delta _j = \gamma_{j,j}.
\end{equation}

Finally, the whole forward propagation of IDN can be defined by
\begin{equation}\label{eq3-9}
	\delta = \mathcal{D} (z; \theta_\gamma ),\ \mathcal{D} = \mathcal{D}_\delta \circ \mathcal{D}_\gamma.
\end{equation}

\subsection{Transfer Learning-Based Framework}\label{Sec3:B}
Transfer learning is an effective learning tool. 
It can assist the learning of the current task (target domain) based on the data or knowledge from another similar task (source domain).
By incorporating prior knowledge from the source domain, transfer learning-based models can achieve superior performance with less training.
They have demonstrated success in industrial scenarios to deal with multi-operation, multi-modal, and multi-block problem. 

In this study, we developed a transfer learning framework,
namely TDN, that leverages normal domain knowledge from a well-trained VAE to aid the IDN in identifying fault signals contained in the faulty samples.
The learning procedure of TDN can be summarized in the following two steps:

\textbf{1) before transfer:} train a VAE to learn fault-free behavior of the system.
The training set $Z^n$ in normal status is divided into several batches
$\{Z^n_{q}\}_{q=1}^{N_q}$, where each batch contains $N_k$ samples denoted as $Z^n_q = \{ z_q^n(k)\}_{k=1}^{N_k}$.
They are then fed into the VAE for forward propagations, represented by
\begin{equation}\label{eq3-10}
    \hat{Z}^n_{q} =  \mathcal{V}(Z^n_{q};\theta _\mathcal{V}),q = 1, 2, ..., N_q,
\end{equation}
where
$\mathcal{V}(\cdot)$ denotes the forward mapping of the VAE defined in (\ref{eq2-2});
$\theta_\mathcal{V} = \{\theta_\mu, \theta_\sigma, \theta_h \}$ represents the parameters of VAE to be optimized. 
Then, the backpropagation algorithm can be utilized to implement end-to-end gradient calculation, by
\begin{equation}\label{eq3-11}
    \nabla_{\theta_\mathcal{V}^{(l)}} = \frac{\partial \mathcal{J}_\mathcal{V}}{\partial \theta_\mathcal{V}^{(l)}} =  \frac{\partial \mathcal{J}_\mathcal{V}}{\partial Z_q^{(L_\mathcal{V})}}  \frac{\partial Z_q^{(L_\mathcal{V})} }{\partial Z_q^{(L_\mathcal{V}-1)}} 
    \cdots  \frac{\partial Z_q^{(l+1)} }{\partial Z_q^{(l)}} \frac{\partial Z_q^{(l)}}{\partial \theta_\mathcal{V}^{(l)}}, 
\end{equation}
where $L_\mathcal{V} $ denotes the number of layers in VAE;
$\mathcal{J}_\mathcal{V} $ is the loss function defined in (\ref{eq2-1-2});
$ Z_q^{(l)}$ and $ \theta_\mathcal{V}^{(l)}$ represent the activation values and parameters in the $(l)^{th}$ layer, respectively.
Root mean square propagation (RMSProp) developed by Hinton \cite{hinton2012neural} can be then employed to update the parameters of the VAE.
where $\varXi_t$ stands for the decaying average over past squared gradients \cite{hinton2012neural} in the $t^{th}$ iteration;
$\eta $ is the learning rate.
\begin{figure*}[!ht]\centering
	\includegraphics[width=14cm]{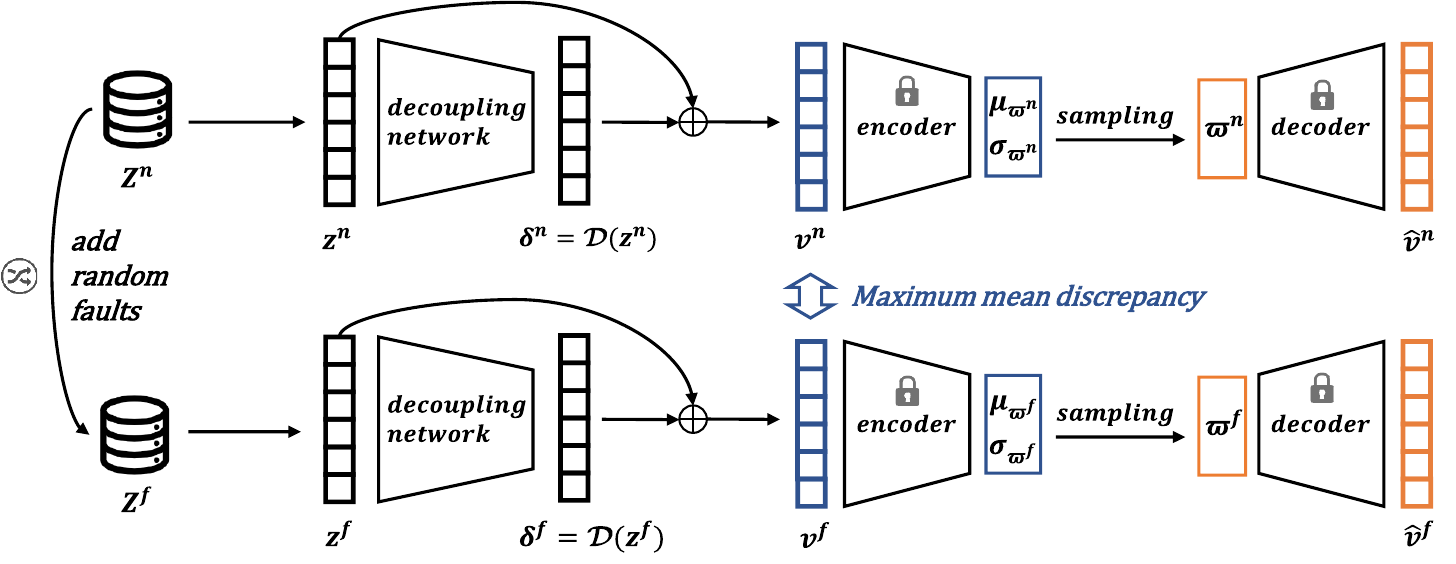}
	\caption{The overall model structure and forward propagation of TDN.}\label{fig3-3}
\end{figure*}

\textbf{2) transfer learning:} train the IDN aided with the well-trained VAE.
\begin{remark}\label{remaark}
The trained VAE can serve as a discriminator to distinguish whether its inputs are in a normal state or not.
In the case of faulty inputs, the loss $ \cal{J}_{\cal{V}}$ will increase significantly compared to the normal inputs \cite{pan2022new}.
To take advantage of this property, we develop a transfer learning-based framework called TDN
to pull the faulty samples back to their normal status with the aid of VAE loss.
\end{remark}
In the transfer training phase, the parameters of VAE $\theta _\mathcal{V}$ are frozen,
which are only used for forward computation and gradient backpropagation but are not involved in updating.
The main objective of this phase is to train the parameters of the IDN $\theta_\gamma$ to identify the
fault signals embedded in the samples via $\mathcal{D}:\mathbb{D}_{z^f}\to  \mathbb{D}_{\tilde{f}}$ and thus to learn the transfer mapping from faulty to
normal $(Z^f\to V^f \in \mathbb{D}_{z^n})$.
With effective training, the IDN and VAE are expected to meet the following goals
\begin{equation}\label{eqfig4}
    \begin{split}
        &\mathcal{D}(Z^n)\simeq 0,\ \mathcal{D}(Z^f)\simeq -f\\
        Z^n&+\mathcal{D}(Z^n)= V^n\simeq \mathcal{V}(V^n) \in \mathbb{D}_{z^n}\\
        Z^f&+\mathcal{D}(Z^f)= V^f\simeq  \mathcal{V}(V^f) \in \mathbb{D}_{z^n},\\
        \end{split}
\end{equation}
where $\mathbb{D}_{z^n}$ denotes the domain of samples in normal status.

As illustrated in Fig. \ref{fig3-3}, the structure and forward propagation of TDN are shown,
which consists of a well-trained VAE and a IDN to be learned.
For a batch-input pair $\{Z_q^n, Z_q^f\}$, the fault-free observations $Z_q^n$ and faulty ones $Z_q^f$ are parallelized and inputted into the IDN.
Then, the sum of inputs and outputs of the IDN $V_q = Z_q + \mathcal{D}(Z_q)$ is fed into the well-trained VAE for obtaining reconstructions.
The forward propagation of TDN can be succinctly described using the following formulas
\begin{equation}\label{eq3-13}
    \begin{split}
    V_q^n = Z_q^n+\mathcal{D}(Z_q^n; \theta_\gamma ), &\ V_q^f = Z_q^f+\mathcal{D}(Z_q^f;\theta_\gamma)\\
    \hat{V}_q^{n}= \mathcal{V} (V_q^n;\theta_\mathcal{V}),&\ \hat{V}_q^{f}= \mathcal{V} (V_q^f;\theta_\mathcal{V}),
	\end{split}
\end{equation}
where $V_q = \{v_q(k)\}_{k=1}^{N_k}$.

Consider using the losses of VAE $\mathcal{J}_\mathcal{V}$ and the loss of
maximum mean discrepancy $\mathcal{J}_{mmd}$ to reach \eqref{eqfig4}.
The loss function of TDN can be then defined by
\begin{equation}\label{eq3-15}
    \begin{split}
    &\mathcal{J} _{tl} = \mathcal{J}_\mathcal{V} (V^n_q)+\mathcal{J}_\mathcal{V} (V^f_q)+ \lambda_{tl} \mathcal{J}_{mmd}(V^n_q,V^f_q)\\
    &\mathcal{J}_{mmd}(V^n_q,V^f_q)={\left\lVert \frac{1}{N_k}\sum_{k = 1}^{N_k}v^n_q(k) - \frac{1}{N_k}\sum_{k = 1}^{N_k}v_q^f(k)\right\rVert}_2^2,
    \end{split}
    \end{equation}
where $\lambda_{tl}$ is a hyperparameter to weigh two parts of the loss;
$\mathcal{J}_{mmd}$ is a widely adopted technique in transfer learning to reduce the distance between two domain centers.
Based on the backpropagation algorithm, $ \partial \mathcal{J}_{tl} /  \partial \theta_\gamma $ is not hard to obtain.
Then, the parameters $\theta_\gamma$ can be updated using the RMSProp algorithm.

It is worth noting that the faulty data used in \eqref{eq3-13} are obtained by adding random fault signals to the normal ones.
This is to prevent the model from being overly sensitive to the specific fault categories and presenting a limited performance
for unseen faults. By introducing random faults, the generalization of the model will be improved.
Concretely, the steps for generating the random faulty dataset in this study can be summarized as follows:
\begin{enumerate}
    \item Add random fault signal $f_{rd,j}$ into variable $j$
    
    \begin{equation}\label{eq3-16}
        z^f = z^n + f_{rd},\ f_{rd,j} = \varsigma_{dire,j} \mathfrak{F}_{amp,j},
    \end{equation}
    where $\varsigma_{dire,j} \sim \mathcal{B}(p_{add})$ determines whether adds a fault in variable $j$;
    $p_{add}$ denotes some probability for adding fault signals;
    $\mathfrak{F}_{amp,j} \sim \mathcal{U}(f_{min,j}, f_{max,j})$;
    $f_{min,j}$ and $f_{max,j}$ are the lower and upper limits of the fault amplitude associated with the real task.

    \item Use the generated fault dataset $Z^f$ and real normal ones $Z^n$ to train the TDN .
    \item Every time the training set $\{Z^n, Z^f\}$ is traversed,
    a new faulty dataset $Z^f$ is generated via step 1 to replace the old one in the training set. Return to step 2 until training is complete..
\end{enumerate}

By feeding both normal and generated faulty data $\{Z^n, Z^f\}$ into the decoupled network, $\{V^n, V^f\}$ can be calculated straightforwardly. 
If $V^n \notin \mathbb{D}_{z^n} $ or $V^f \notin \mathbb{D}_{z^n}$,
then the VAE will receive a significant reconstruction error \cite{pan2022new} driving parameter updates for IDN.
In this way, $V^n$ and $V^f$ will be both dragged into the fault-free domain with a small maximum mean discrepancy.
After training, the IDN can separate the fault signal contained in the observations effectively.

In summary, the transfer procedure of TDN can be described by Algorithm 1.
\begin{algorithm}[!bht]
    \SetKwInOut{Input}{Input}
    \SetKwInOut{Output}{Output}
    \caption{Transfer Learning phase for TDN}\label{alg1}
    \Indentp{-0.5em}
    \Input{Training set $Z^n$ in normal status, number of iterations $N_e$, number of batches $N_q$, trained VAE.}
    \Output{Well-trained TDN.}
    \Indentp{0.5em}
    \BlankLine
    Standardize $Z^n$ to zero mean and unit variance\;
    Freeze the parameters $\theta_{\mathcal{V}}$ in VAE \;
    \For{$e\leftarrow 1$ \textbf{\textup{to}} $N_e$}
        {
            Generate a new random faulty dataset $Z^f$ via (\ref{eq3-16})\;
            
            \For{$q\leftarrow 1$ \textbf{\textup{to}} $N_q$}
            {
                Draw a batch of data from $\{Z^n, Z^f\}$, denoted as $\{Z_q^n, Z_q^f\}$\;
                Compute $\{ \hat V_q^n, \hat V_q^f \}$ via (\ref{eq3-13})\;
                Calculate loss via (\ref{eq3-15}) and then update $\theta_\gamma$ using RMSProp algorithm.
            }
            
        }
\end{algorithm}

\subsection{TDN-Based Fault Detection and Estimation}\label{Sec3:C}

After the TDN is well-trained, it can be employed for FD and FE purposes.
Unlike fault classifiers that are sensitive only to familiar and known fault categories,
FD approaches aim to identify any intolerable deviations in system behavior caused by faults through the constructed residual generators.
Typically, FD methods are designed in two steps: residual generation and residual evaluation.
Residuals can be generated by trained systematic observers or by models fitted to the observed data.
A threshold is then learned to characterize the bound of the residuals generated from the offline normal data,
which also serves as a comparison criterion for subsequent online monitoring.

The generated residuals should satisfy the basic FD condition (\ref{eq2-7}).
Based on (\ref{eqfig4}), one can deduce that the well-trained IDN can be applied to generate residuals. We can establish the IDN-based residual generator as
\begin{equation}\label{eq3-23}
   \phi (z) =\mathcal{D}(z).
\end{equation}

After residual generation, $T^2(\phi(z))$ test statistics is selected for converting the residuals into single values.
It can be defined by
\begin{equation}\label{eq3-24}
    T^2(\phi(z)) = (\phi(z))^T \varSigma ^{-1} _ {\phi(z^n)} \phi(z),
\end{equation}
where 
\begin{equation}\label{eq3-25}
    \varSigma _ {\phi(z^n)} = \frac{1}{N_k} \sum _{k=1}^{N_k} \phi(z^n) (\phi(z^n))^T.
\end{equation}
$\varSigma _ {\phi(z^n)}$ represents the covariance matrix of the collected normal observations.
$z$ in (\ref{eq3-24}) can be $z^n$ or $z^o$, representing offline and online samples, respectively.
$\phi(z)$ in (\ref{eq3-24}) and (\ref{eq3-25}) are centered before computation.

\begin{remark}\label{remark3}
The IDN proposed in this paper is input-output decoupled and the residual variables generated by it are uncorrelated with each other. 
Therefore, their covariance matrix and inverse will be approximated as a unit array ($\varSigma \simeq \varSigma^{-1} \simeq I$).
In this case, $T^2$ is equivalent to the sum of squares of the residual variables. 
\end{remark}
Threshold setting is the subsequent operation to test statistical calculation.
In general, if the distribution of test statistics is accessible, the inverse function of the cumulative density distribution can be applied directly for threshold setting.
Instead, a beforehand probability density function (PDF) estimation program is required when the PDF of test statistics is unknown.
In this paper, kernel density estimation (KDE) is applied to estimate PDF.
The detailed description of KDE and threshold setting procedure can be found in \cite{kde}.

In summary,
the whole training process of TDN for the FD purpose can be described by four step: 
1) pre-train VAE via $Z^n$;
2) train TDN via $Z^n$ and generated $Z^f$ by Algorithm 1;
3) generate residuals from TDN and calculate $T^2$ by \eqref{eq3-23} and \eqref{eq3-24};
4) estimate PDF and set the threshold $J_{th}$ via KDE \cite{kde}.
Then, for an online sample $z^o$, its state can be determined by
\begin{equation}
\begin{cases}
    z^o\rm{\ in\ normal\ status}, & \text{ if } T^2(\phi(z^o)) \le J_{th} \\
    z^o\rm{\ in\ faulty\ status}, & \text{ if } T^2(\phi(z^o)) > J_{th}
\end{cases}
\end{equation}

In addition to FD purposes, the developed TDN is also capable of FE tasks.
According to (\ref{eqfig4}), one can know that the trained IDN can act as a fault signal extractor. 
This means the fault signal can be predicted directly using the output of the IDN by
\begin{equation}\label{eq3-34}
    \varphi  (z) = -\mathcal{D}(z) = \tilde{f} \simeq f.
\end{equation}
Since the output variables of IDN are uncorrelated with each other,
the prediction of $f_j$ is not affected by the other faults $f_i$, $i \ne j$, which is favorable for the prediction accuracy.

In order to evaluate the FD performance of the proposed method, false alarm rate (FAR) and missed detection rate (MDR) are applied in this study.
Consider that testing set $Z^o = \{ Z^o_{(c)} \}_{c=1}^{N_c}$ consists of multiple fault-type datasets collected from different simulation runs.
Each of run produces one fault-type data, in which the first few observations are in normal status, and
the faults are introduced after a period of time.
For the $(c)^{th}$ type of faults, the following indicators can be defined to calculate the FAR and MDR
\begin{equation}\label{eq3-32}
    \begin{split}
      p_{FAR,(c)}&= N_{FA,(c)}/ \left(N_{FA,(c)}+N_{TA,(c)}\right)\\
      p_{MDR,(c)}&= N_{MD,(c)}/ \left(N_{MD,(c)}+N_{RD,(c)}\right),
    \end{split} 
\end{equation}
where $FA$, $TA$, $MD$, and $RD$ represent false alarm, true alarm, missed detection, and right detection, respectively.
Their meanings are detailed in  \ref{tab1}.
Similarly, the average FAR (AFAR) and average MDR (AMDR) can be defined to test the overall FD performance.
\begin{table}[!htb]\centering
    \caption{Explanation of $FA$, $TA$, $MD$, and $RD$}\label{tab1}
    \begin{tabular}{p{3.0cm}p{2.0cm}p{2.0cm}}
    \toprule[0.8pt]
    \makecell[c]{Condition description}  &  \makecell[c]{$f=0$} & \makecell[c]{$f \ne 0$} \\
    \hline\hline \noalign{\smallskip}
    \makecell[c]{$T^2(\phi (z)) \le J_{th}$} & \makecell[c]{$RD$} & \makecell[c]{$MD$} \\
    \makecell[c]{$T^2(\phi (z)) > J_{th}$} & \makecell[c]{$FA$} & \makecell[c]{$TA$}
    \cr
    \hline
    \end{tabular}
\end{table}

In addition, the root mean square error (RMSE) can be adopted to evaluate the FE performance of the developed method,
for the $(c)^{th}$ fault type, which can be defined by
\begin{equation}\label{eq3-35}
    {RMSE}_{(c)} =\sqrt{ \frac{1}{N_{k,(c)}} \sum_{k=1}^{N_{k,(c)}} \left\lVert \tilde f(k)-f(k) \right\rVert^2_2 }, 
\end{equation}
where $N_{k,(c)}$ represents the number of samples in the $(c)^{th}$ faulty dataset.
Then, the average RMSE (ARMSE) can be defined accordingly.

\renewcommand\arraystretch{1.5}
\begin{table}[b]
    \centering
    \caption{Description of faults in the numerical example}
    \label{fault_num}
    \resizebox{0.5\textwidth}{!}{%
    \begin{tabular}{cl}
        \toprule[0.8pt]
        \hline
    Fault Type & Description \\ 
    \hline
    Fault01 & $v_{1}(k)\sim \mathcal{N}(\textcolor{red}{0.2},0.01)$ \\
    Fault02&$\begin{aligned} z_{2}(k)&=\textcolor{red}{0.5z_2(k-1)}+{\rm sin}(z_2(k-1)+\left(z_{2}(k-1)\right)^2
        \\&+2 \left(z_{2}(k-1)\right)^3) +v_{2}(k)\end{aligned} $ \\
    Fault03 & simultaneous Fault01 and Fault02 \\
    Fault04&  $z_{1}(k)={\rm sin}(z_{1}(k-1)) +\textcolor{red}{ 0.3} e^{{\rm cos}(z_{1}(k-1))} + v_{1}(k)$\\
    Fault05& $x_{1}(k) = x_{1}(k) + \textcolor{red}{0.0018(k-199)}$\\
    Fault06& $x_{2}(k) = x_{2}(k) + \textcolor{red}{0.005(k-199)}$ \\ 
    Fault07& $x_{3}(k) = x_{3}(k) + \textcolor{red}{\vert 0.4{\rm sin}(\frac{2\pi(k-199)}{300}+\frac{\pi}{22})\vert }$\\ 
    Fault08& $x_{4}(k) = x_{4}(k) + \textcolor{red}{0.009(k-199)}$\\ 
    Fault09& $x_{4}(k) = x_{4}(k) + \textcolor{red}{1.8{\rm sin}(\frac{\pi(k-199)\%300}{600}+\frac{\pi}{33})}$\\ 
    Fault10& $x_{5}(k) = x_{5}(k) - \textcolor{red}{0.01(k-199)}$ \\ 
     \toprule[0.8pt]
     \hline
    \end{tabular}%
    }
    \end{table}
\section{Experimental Verification }\label{Sec4}
\renewcommand\arraystretch{1.5}
\begin{table*}[t]
    \centering
    \caption{Selected 3 $\times$ 6 TDNs with different network architectures ($\mathcal{D}$s for IDNs and $\mathcal{V}$s for VAEs)}
    \label{structures}
    \begin{tabular}{cc}
        \toprule[0.8pt]
    Structure&Detailed network structures of $\mathcal{D}$s and $\mathcal{V}$s for constructing TDNs (input layer on the right-hand side)\\
    \hline
    \hline
    $\mathcal{D}_1$ & $ \mathcal{A} _{A}\circ\mathcal{L}[5,100] \circ  \mathcal{A} _{A} \circ \mathcal{L}[100,100] \circ \mathcal{A} _{A} \circ  \mathcal{L}[100,5]$ \\ 
    $\mathcal{D}_2$ & $ \mathcal{A} _{A}\circ\mathcal{L}[5,100] \circ  \mathcal{A} _{Q} \circ \mathcal{L}[100,100] \circ \mathcal{A} _{A} \circ  \mathcal{L}[100,5]$ \\ 
    $\mathcal{D}_3$ & $ \mathcal{A} _{A}\circ\mathcal{L}[5,100] \circ  \mathcal{A} _{G} \circ \mathcal{L}[100,100] \circ \mathcal{A} _{A} \circ  \mathcal{L}[100,5]$ \\ 
    $\mathcal{V}_1$ &  $ \mathcal{A} _{A}\circ\mathcal{L}[5,100] \circ  \mathcal{A} _{A} \circ \mathcal{L}[100,20]\circ\mathcal{P}_S \circ \{\mathcal{A} _{A}\circ\mathcal{L}[10,20],\mathcal{A} _{A}\circ\mathcal{L}[10,20]\} \circ \mathcal{A} _{A} \circ \mathcal{L}[100,5]$ \\ 
    $\mathcal{V}_2$ &  $ \mathcal{A} _{A}\circ\mathcal{L}[5,100] \circ  \mathcal{A} _{T} \circ \mathcal{L}[100,20]\circ\mathcal{P}_S \circ \{\mathcal{A} _{A}\circ\mathcal{L}[10,20],\mathcal{A} _{A}\circ\mathcal{L}[10,20]\}  \circ \mathcal{A} _{A} \circ \mathcal{L}[100,5]$ \\ 
    $\mathcal{V}_3$ &  $ \mathcal{A} _{A}\circ\mathcal{L}[5,100] \circ  \mathcal{A} _{A} \circ \mathcal{L}[100,20]\circ\mathcal{P}_S \circ \{\mathcal{A} _{A}\circ\mathcal{L}[10,20],\mathcal{A} _{A}\circ\mathcal{L}[10,20]\}  \circ \mathcal{A} _{T} \circ \mathcal{L}[100,5]$ \\ 
    $\mathcal{V}_4$ &  $ \mathcal{A} _{A}\circ\mathcal{L}[5,100] \circ  \mathcal{A} _{S} \circ \mathcal{L}[100,20] \circ  \mathcal{A} _{S} \circ \mathcal{L}[20,10]\circ\mathcal{P}_S \circ  \{\mathcal{A} _{A}\circ\mathcal{L}[10,20],\mathcal{A} _{A}\circ\mathcal{L}[10,20]\} \circ\mathcal{A} _{S} \circ \mathcal{L}[20,100]\circ \mathcal{A} _{G} \circ \mathcal{L}[100,5]$ \\ 
    $\mathcal{V}_5$ &  $ \mathcal{A} _{A}\circ\mathcal{L}[5,100] \circ  \mathcal{A} _{S} \circ \mathcal{L}[100,20] \circ  \mathcal{A} _{S} \circ \mathcal{L}[20,10]\circ\mathcal{P}_S \circ \{\mathcal{A} _{A}\circ\mathcal{L}[10,20],\mathcal{A} _{A}\circ\mathcal{L}[10,20]\} \circ\mathcal{A} _{S} \circ \mathcal{L}[20,100]\circ \mathcal{A} _{Q} \circ \mathcal{L}[100,5]$ \\ 
    $\mathcal{V}_6$ &  $ \mathcal{A} _{A}\circ\mathcal{L}[5,100] \circ  \mathcal{A} _{S} \circ \mathcal{L}[100,20] \circ  \mathcal{A} _{S} \circ \mathcal{L}[20,10]\circ\mathcal{P}_S \circ \{\mathcal{A} _{A}\circ\mathcal{L}[10,20],\mathcal{A} _{A}\circ\mathcal{L}[10,20]\} \circ\mathcal{A} _{A} \circ \mathcal{L}[20,100]\circ \mathcal{A} _{Q} \circ \mathcal{L}[100,5]$ \\ 
     \hline
     \toprule[0.8pt]
    \end{tabular}
    \end{table*}
    \renewcommand{\arraystretch}{1.2}
\begin{table*}[!htb]\centering
\caption{FD comparison results of TDNs and other four advanced methods on the numerical example}\label{num_result}
\begin{tabular}{ccccccccccccccc}
\toprule
\multirow{3}{*}{Structure} & \multicolumn{7}{c}{The mean of $\bar p_{FAR}$} & \multicolumn{7}{c}{The mean of $\bar p_{MDR}$} \\ 
\cmidrule(r){2-8} \cmidrule(l){9-15} 
 & \multirow{2}{*}{VAE} & \multirow{2}{*}{K-VAE} & \multirow{2}{*}{AAE} & \multirow{2}{*}{DAE}
 & \multicolumn{3}{c}{TDN} & \multirow{2}{*}{VAE} & \multirow{2}{*}{K-VAE} & \multirow{2}{*}{AAE} & \multirow{2}{*}{DAE} 
 & \multicolumn{3}{c}{TDN} \\ 
 \cmidrule(r){6-8} \cmidrule(l){13-15}
 & & & & &$\mathcal{D}_1$&$\mathcal{D}_2$&$\mathcal{D}_3$& & & & &$\mathcal{D}_1$&$\mathcal{D}_2$&$\mathcal{D}_3$\\
\hline\hline \noalign{\smallskip}
$\mathcal{V}_1$ & 0.30 & 0.95 & 0.45 & 0.40 & 0.40 & 0.40 & 0.45 & 10.73 & 16.02 & 7.56 & 7.56 & 7.56 & \textbf{7.55} & 9.55 \\
$\mathcal{V}_2$ & 0.25 & 0.85 & 0.50 & 0.65 & 0.40 & 0.45 & 0.40 & 9.88 & 8.86 & 8.8 & 7.61 &\textbf{7.56} &\textbf{7.56} & 9.6 \\
$\mathcal{V}_3$ & 0.30 & 0.75 & 0.60 & 0.50 & 0.40 & 0.40 & 0.40 & 9.71 & 8.51 & 7.62 & 7.63 & 7.56 & \textbf{7.55} & 9.56 \\
$\mathcal{V}_4$ & 0.75 & 0.55 & 0.60 & 0.50 & 0.40 & 0.40 & 0.45 & 7.96 & 8.23 & 9.06 & 7.61 & 7.56 & \textbf{7.53} & 9.72 \\
$\mathcal{V}_5$ & 0.8 & 0.55 & 0.35 & 0.45 & 0.40 & 0.40 & 0.45 & 7.95 & 8.31 & 8.13 & 7.76 & \textbf{7.56} & \textbf{7.56} &9.61 \\
$\mathcal{V}_6$ & 0.65 & 0.40 & 0.45 & 0.45 & 0.40 & 0.40 & 0.40 & 8.00 & 8.93 & 7.86 & 7.61 & \textbf{7.56} & 7.60 & 9.60\\
\hline
\end{tabular}
\end{table*}

In this section, a numerical example and a TTS simulation are used to verify the effectiveness of the proposed TDN model.
The simulation results illustrate that the TDN has superior FD and FE performance than other advanced approaches.
\subsection{Numerical example}\label{Sce4:A}
The nonlinear dynamics of the selected numerical example can be defined as
\begin{equation}\label{eq_numerical}
\begin{aligned}
z_{1} &= 0.1 x_{1} + x_{1}/\sqrt{\left(x_{1} \right)^2 + \left(x_{2}\right)^2} + w_{1}\\
z_{2} &= 0.1 x_{1}  x_{2} + x_{2}/\sqrt{\left(x_{1} \right)^2 + \left(x_{2}\right)^2} + w_{2}\\
z_{3} &= \cos(x_{1})^3 + 0.1 e^ {{\rm sin}(x_{2})} + w_{3}\\
z_{4} &= {\rm sin}(x_{1})^3 + \log(2+{\rm cos}(x_{2}))+ w_{4}\\
z_{5} &= \sqrt{\left(x_{1} \right)^2 + \left(x_{2}\right)^2} + 0.1  \left(x_{1} \right)^3 -0.1 \left(x_{1} \right)^4 \\
&+ {\rm sin}(0.1 x_{1} x_{2})  + w_{5},
\end{aligned}
\end{equation}
where the timestamps $(k)$ of the observations $z_1$ to $z_5$ are omitted for simplicity;
$w_{1},w_{2},w_{3},w_{4},w_{5}$ are Gaussian noises with zero mean and standard deviations, respectively,
set as $0.05,0.16,0.02,0.05,0.3$; $x_{1}$ and $x_{2}$ are latent variables that are unobservable, satisfying
\begin{equation}\label{eq_numerical_lv}
\begin{aligned}
    x_{1}&={\rm sin}(x_{1}^{\dagger}) + 0.1 e^{{\rm cos}(x_{1}^{\dagger})} + v_{1}\\
    x_2&={\rm sin}(x_2^{\dagger}+\left(x_{2}^{\dagger}\right)^2+2 \left(x_{2}^{\dagger}\right)^3) +v_{2},
\end{aligned}
\end{equation}
where the timestamps $(k)$ of the variables are omitted and $(k-1)$ are replaced by superscript ``$\dagger$ '' for simplicity;
$v_{1}$ and $v_{2}$ are Gaussian noises with zero mean and standard deviations of $0.01$.

A total of 15000 normal samples were generated for offline training and 1000 $\times$ 10 samples (including 200 $\times$ 10 normal samples and 800 $\times$ 10 faulty samples) were collected for online monitoring.
It means that the online datasets were collected from ten different simulations, with the first 200 of each simulation being normal samples and the last 800 being faulty samples.
Each online dataset simulated a category of faults, and their descriptions can be found in TABLE \ref{fault_num},
in which red parts highlightthe location of the faults; \% denotes the remainder operation.

In order to investigate the performance of the proposed method,
three different network structures of $\mathcal{D}$ and six different $\mathcal{V}$ were constructed in TABLE \ref{structures}, 
where $\mathcal{D}_1$ to $\mathcal{D}_3$ indicate the structure of different IDNs, while $\mathcal{V}_1$ to $\mathcal{V}_6$ represent the structure of VAEs;
$\mathcal{L}[m_{(l)}, m_{(l-1)}]$ denotes the linear mapping layer given in \eqref{eq3-6} with the number of input and output variables, respectively,
as $m_{(l-1)}$ and $ m_{(l)}$; 
$\mathcal{P}_S$ represents sampling operation defined in \eqref{eq2-2};
the subscripts of $\mathcal{A}$ represent different types of activation functions, defined by
\begin{equation}\label{eq5-6}
    \begin{split}
        \textrm{Affine:}&\ \ \mathcal{A}_A(\iota)  = \iota\\
        \textrm{Square:}&\ \ \mathcal{A}_Q(\iota)  = \iota ^2\\
        \textrm{Sigmoid:}&\ \ \mathcal{A}_S(\iota)  = 1 / (1 + {e^{ -\iota }}) \\
        \textrm{Gaussian:}&\ \ \mathcal{A}_G(\iota )  = 1 - {e^{ - {\iota ^2}}} \\
        \textrm{Tanh:}&\ \ \mathcal{A}_T(\iota)  = ({e^\iota } - {e^{ - \iota }})/({e^\iota} + {e^{ - \iota}}).
    \end{split}
\end{equation}

Then, 18 different TDNs are constructed by pairing 3 IDNs and 6 VAEs defined in TABLE \ref{structures}.
Moreover, DAE \cite{pan2022new}, VAE \cite{pan2022new}, AAE \cite{AAE_hanguo} and K-VAE \cite{kaiwang_VAE} are used as baseline methods to demonstrate the effective of the proposed method,
all of which adopt $\mathcal{V}_1 $ to $ \mathcal{V}_6$ as their network structures in comparative verification.
In addition to the network structure, other hyperparameters of the five-type models are selected as follows under a lot of trial and error:
the number of training equals 15; batch size is set as 16; 
learning rate $\eta  = 0.001$; $\lambda _{tl}$ in \eqref{eq3-15} is set as 0.1. 
Besides, the expect FAR $E[\hat p_{FAR}]$ takes the value of $ 0.5\%$.

Based on the different network structures selected in TABLE \ref{structures},
a large number of experiments were conducted in the numerical examples.
For each model, ten independent repetitions were performed to eliminate the effect of randomness.
As shown in TABLE \ref{num_result}, 
the means of AFAR and AMDR for ten trials are given.

\begin{figure}[t]\centering
    \begin{subfigure}{.5\textwidth}
        \centering
        \includegraphics[width=.88\linewidth]{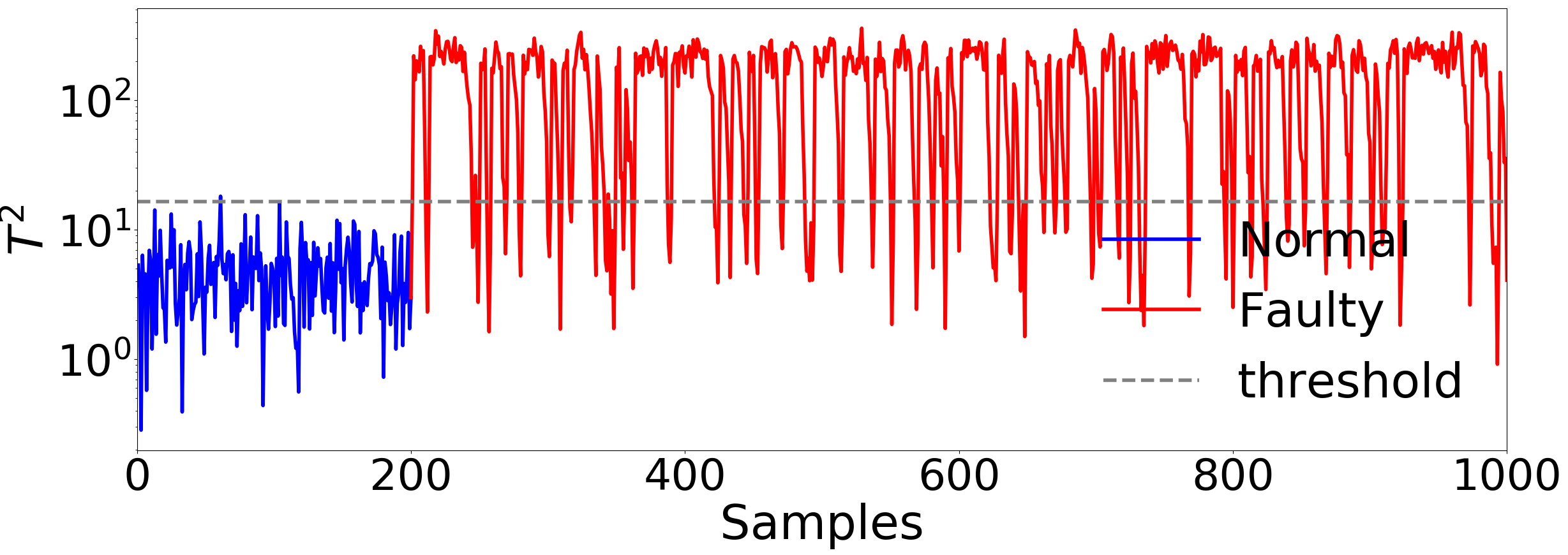}  
        \caption{$Fault\;02$}
        \label{fignum-a}
    \end{subfigure}
    \newline
    \begin{subfigure}{.5\textwidth}
        \centering
        \includegraphics[width=.88\linewidth]{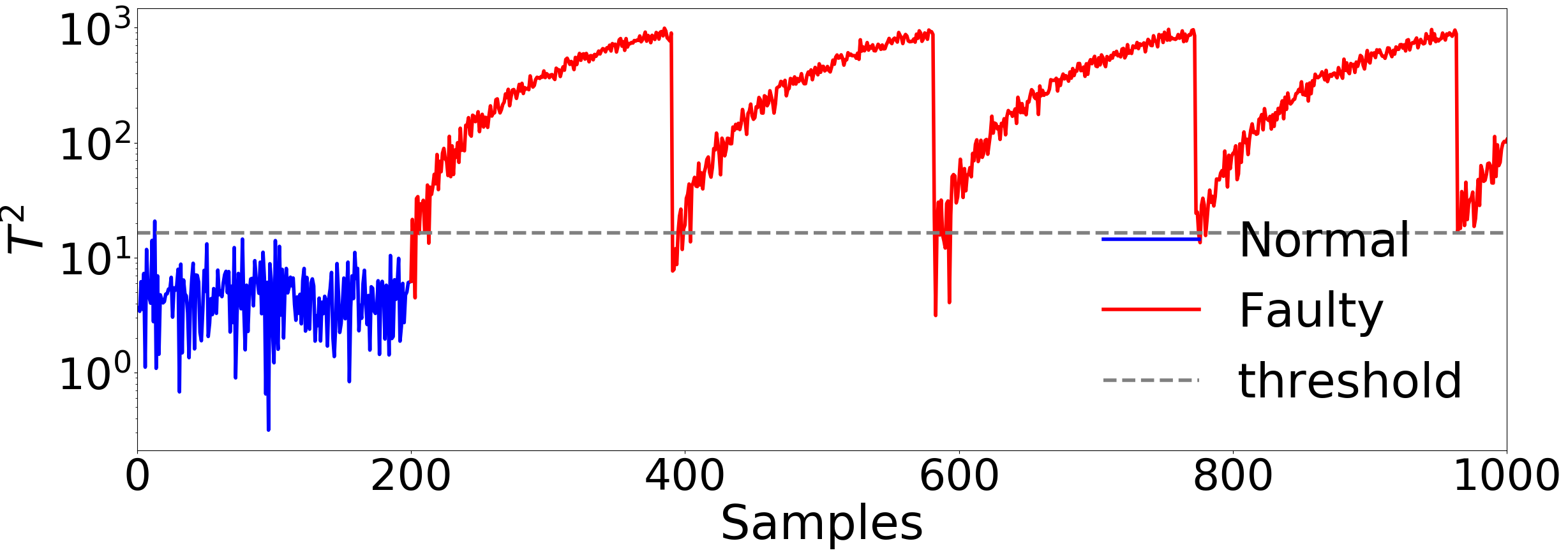}  
        \caption{$Fault\;09$}
        \label{fignum-b}
    \end{subfigure}
    \caption{FD curves of TDN-$\mathcal{D}_2$-$\mathcal{V}_4$ on the numerical example}
    \label{fignum}
\end{figure}

The AFAR of the four methods except TDN appears to be excessively high for specific activation function combinations,
which has exceeded the allowable values. 
In contrast, the AFARs provided by the TDNs are all constraint-satisfying. In terms of AMDR,
TDN-$\mathcal{D}_1$ and TDN-$\mathcal{D}_2$ achieve the best FD performance under different $\mathcal{V}$-structures,
with the AFAR no higher than $0.50\%$ and the lowest AMDR mong the selected models.
Among them, the best result is obtained by TDN-$\mathcal{D}_2$-$\mathcal{V}_4$, taking the lowest AMDR of 7.53\%.
In addition, TDN-$\mathcal{D}_3$ exhibits a relatively higher AMDR than -$\mathcal{D}_1$ and -$\mathcal{D}_2$,
which reveals that the structure selection of $\mathcal{D}$ has a significant effect on the FD performance of TDNs.
``Affine'' and ``Square'' activation functions are more suitable for $\mathcal{D}$
in TDN compared with ``Gaussian'' activation function.
Instead, $\mathcal{V}$ exhibits light influence on FD performance of TDN, but 
VAE, AAE and K-VAE exhibit instability in their FD results with the structural change of $\mathcal{V}$.
Intuitively, Fig. \ref{fignum} displays the results of using TDN-$\mathcal{D}_2$-$\mathcal{V}_4$ to detect Fault 02 and 09 as examples.
The blue and red lines refer to the genuine labels of the test samples being normal or faulty, respectively.
It can be seen that the dashed line (threshold line) can correctly distinguish most of the normal and abnormal samples.

\begin{figure}[t]\centering
	\includegraphics[width=.85\linewidth]{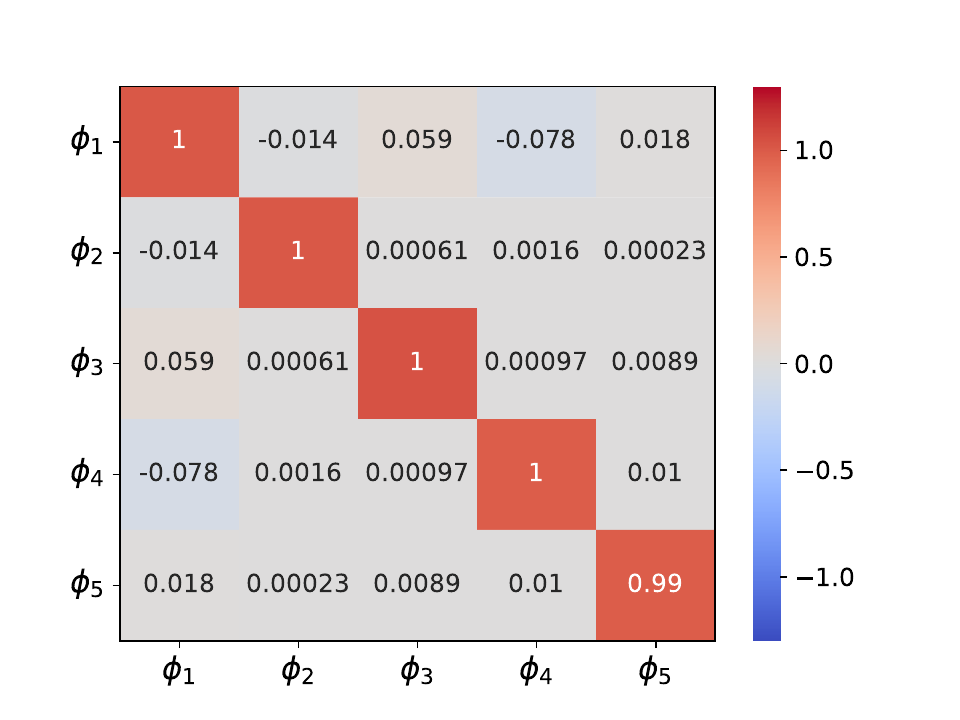}
	\caption{Covariance matrix of TDN-$\mathcal{D}_2$-$\mathcal{V}_4$-generated $\phi(z^n)$}\label{fig_cov_num}
\end{figure}

Moreover, in order to demonstrate the decoupling capabilities of IDNs or TDNs, the covariance matrix of $\phi (z^n)$ generated by TDN-$\mathcal{D}_2$-$\mathcal{V}_4$ is shown in Fig. \ref{fig_cov_num},
$\phi _ j$ is the $j^{th}$ component of $\phi ( z ^ n ), j = 1,2,..., m _ \phi$.
One can conclude that except for the diagonal elements, the values at other positions approach zero.
As noted in Remark \ref{remark3}, the matrix is close to a unit array,
which helps in locating the fault variables and obtaining more accurate FE results.

\renewcommand{\arraystretch}{1.2}
\begin{table}[!htb]\centering
    \caption{FE Results (ARMSEs) given by TDNs}\label{esti_num}
    \begin{tabular}{>{\centering\arraybackslash}p{1.6cm}>{\centering\arraybackslash}p{1.6cm}>{\centering\arraybackslash}p{1.6cm}>{\centering\arraybackslash}p{1.6cm}}
    \toprule
    \multirow{2}{*}{\makecell[c]{Structure}} & \multicolumn{3}{c}{TDN (IDN+VAE)}  \\
    \cmidrule(r){2-4} 
     & $\mathcal{D}_1$ & $\mathcal{D}_2$ & $\mathcal{D}_3$ \\
    \hline\hline \noalign{\smallskip}
    $\mathcal{V}_1$ & 0.994 & 1.145 & 9.426 \\
    $\mathcal{V}_2$ & 0.998 & 1.063& 9.354  \\
    $\mathcal{V}_3$ & 0.999 & 1.352& 9.380\\
    $\mathcal{V}_4$ & 0.993 & 1.618 & 9.383 \\
    $\mathcal{V}_5$ & 1.024 & 1.268 & 9.341 \\
    $\mathcal{V}_6$ & 0.996 & 2.037 & 9.406 \\
    \hline
    \end{tabular}
\end{table}
\renewcommand{\arraystretch}{1.2}
\begin{table}[!htb]\centering
    \caption{FE Results (ARMSEs) given by TL-FCNN}\label{fcnn_esti_num}
    \begin{tabular}{>{\centering\arraybackslash}p{1.6cm}>{\centering\arraybackslash}p{1.6cm}>{\centering\arraybackslash}p{1.6cm}>{\centering\arraybackslash}p{1.6cm}}
    \toprule
    \multirow{2}{*}{\makecell[c]{Structure}} & \multicolumn{3}{c}{TL-FCNN (FCNN+VAE)}  \\
    \cmidrule(r){2-4} 
     & $\mathcal{D}_1$ & $\mathcal{D}_2$ & $\mathcal{D}_3$ \\
    \hline\hline \noalign{\smallskip}
    $\mathcal{V}_1$ & 2.907 & 32.21& 10.11 \\
    $\mathcal{V}_2$ & 2.9551 & 32.66& 9.994  \\
    $\mathcal{V}_3$ & 2.9343 & 32.50& 9.954\\
    $\mathcal{V}_4$ & 3.8437 & 56.02 & 10.24 \\
    $\mathcal{V}_5$ & 3.9966 & 56.46 & 10.23 \\
    $\mathcal{V}_6$ & 3.9353 & 57.44 & 10.24 \\
    \hline
    \end{tabular}
\end{table}

\begin{figure}[!htb]\centering
    \begin{subfigure}{.5\textwidth}
        \centering
        \includegraphics[width=.88\linewidth]{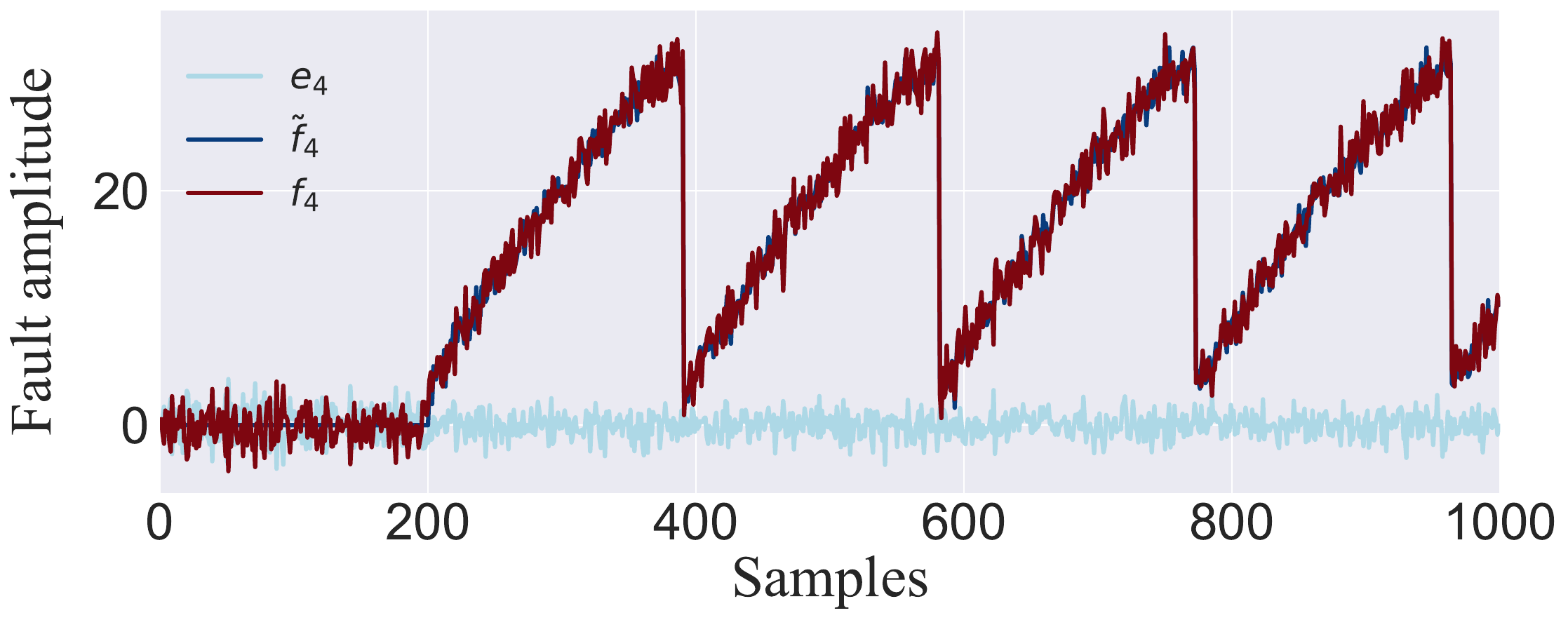}  
        \caption{Fault09 estimated by {TDN}-$\mathcal{D}_1$-$\mathcal{V}_4$ (Affine)}
        \label{fig_esti-c}
    \end{subfigure}
    \newline
    \begin{subfigure}{.5\textwidth}
        \centering
        \includegraphics[width=.88\linewidth]{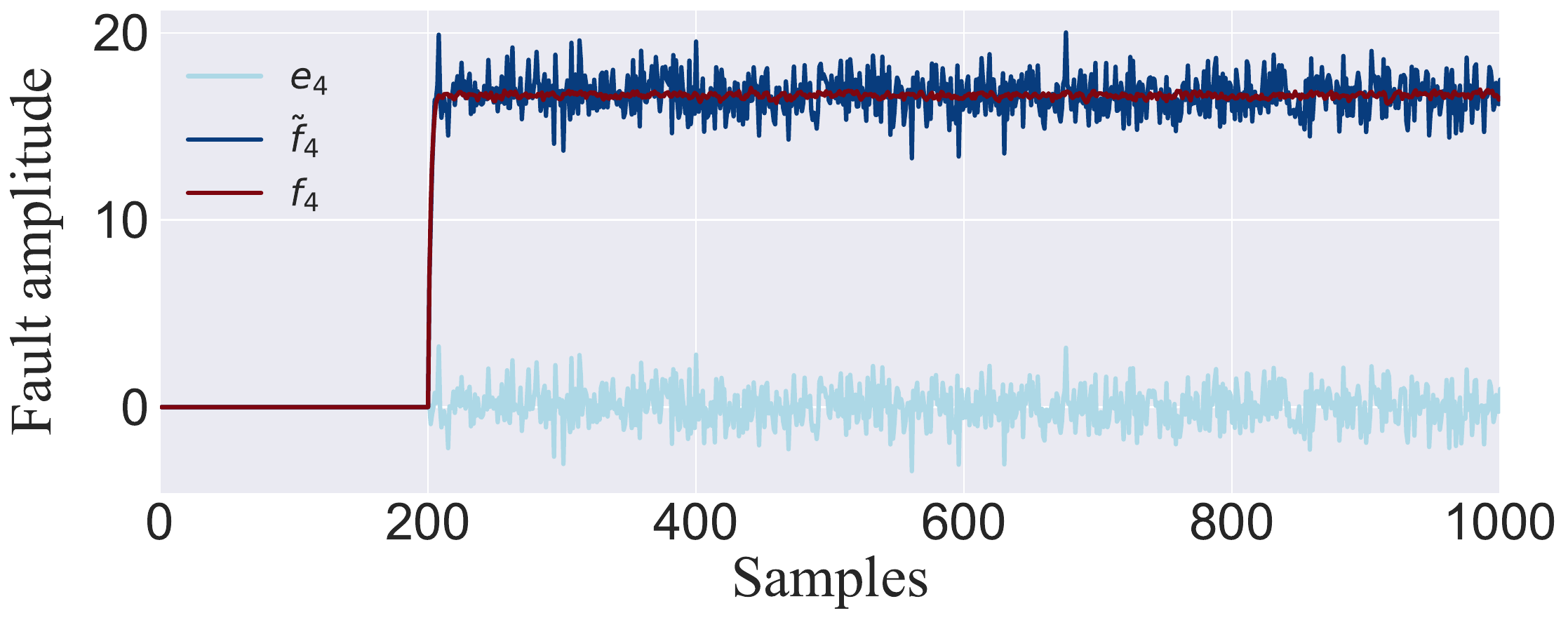}  
        \caption{Fault01 estimated by {TDN}-$\mathcal{D}_1$-$\mathcal{V}_4$ (Affine)}
        \label{fig_esti-a}
    \end{subfigure}
    \newline
    \begin{subfigure}{.5\textwidth}
        \centering
        \includegraphics[width=.88\linewidth]{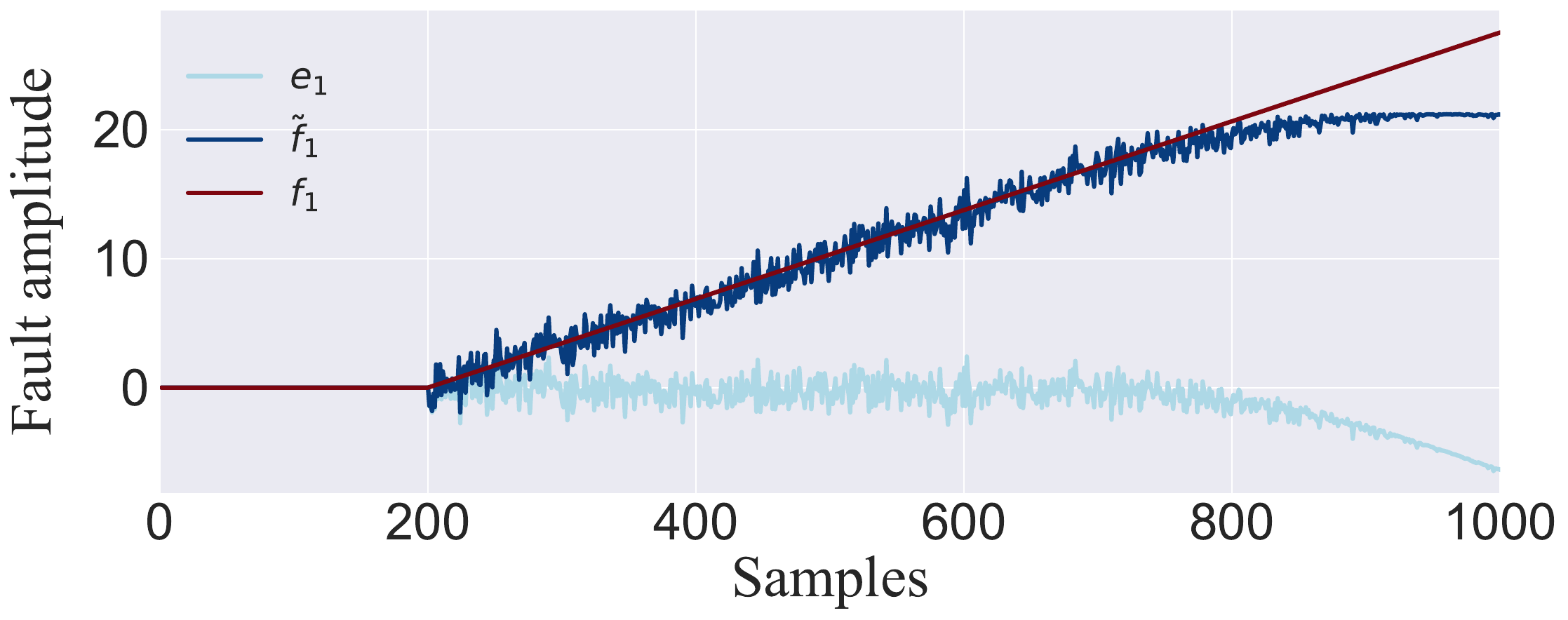}  
        \caption{Fault01 estimated by {TDN}-$\mathcal{D}_3$-$\mathcal{V}_4$ (Gaussian)}
        \label{fig_esti-b}
    \end{subfigure}
    \begin{subfigure}{.5\textwidth}
        \centering
        \includegraphics[width=.88\linewidth]{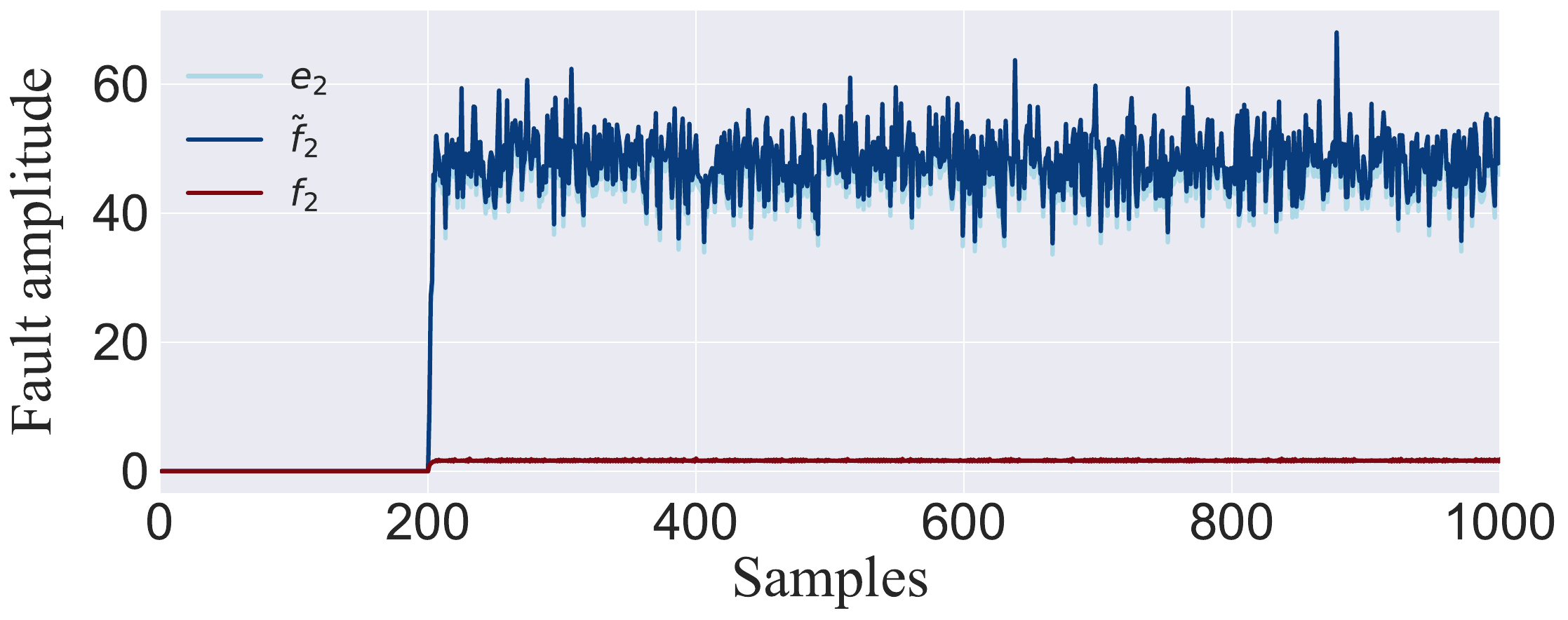}  
        \caption{Fault01 estimated by {TL-FCNN}-$\mathcal{D}_2$-$\mathcal{V}_4$ (Square)}
        \label{fig_esti-b}
    \end{subfigure}
    \caption{
        FE curves on the numerical example given by TDN-$\mathcal{D}_1$-$\mathcal{V}_4$, TDN-$\mathcal{D}_3$-$\mathcal{V}_4$ and 
        TL-FCNN-$\mathcal{D}_2$-$\mathcal{V}_4$}
    \label{fig_esti}
    \vspace{-20pt}
\end{figure}

To illustrate the positive effect of decoupling on the FE, the comparison results of TDNs (IDN+VAE) and TL-FCNNs (FCNN+VAE) on the numerical example are shown in TABLE \ref{esti_num} and TABLE \ref{fcnn_esti_num}.
Here, TL-FCNNs are obtained by replacing IDN in TDN with FCNN, which also still use the TL-based training framework proposed in Section \ref{Sec3}.
The ARMSE is used to evaluate the FE performance of TDNs and TL-FCNNs.
Comparing TABLE \ref{esti_num} and TABLE \ref{fcnn_esti_num},
one can see that the proposed TDNs can achieve more satisfactory FE effectiveness than TL-FCNNs.
As presented in TABLE \ref{esti_num}, most of the FE results for TDNs are excellent,
except for the large estimation errors obtained by TDN-$\mathcal{D}_3$s.
This scenario is similar to the previously discussed FD results,
being influenced more by the structure of $\mathcal{D}$ and less by the structure of $\mathcal{V}$.
In the selected structures, TDN-$\mathcal{D}_1$-$\mathcal{V}_4$ gets the lowest ARMSE 0.993.
Contrastingly, the ARMSEs provided by TL-FCNNs is universally larger than that of TDNs,
and especially the results under the $\mathcal{D}_2$ structure are unacceptable,
reaching a maximum of 57.44. These results provide ample evidence of the assistance of decoupling in improving FE accuracy.

Intuitively, Fig. \ref{fig_esti} displays the representative estimation curves for some faults using {TDN}-$\mathcal{D}_1$-$\mathcal{V}_4$, {TDN}-$\mathcal{D}_3$-$\mathcal{V}_4$,
and TL-FCNN-$\mathcal{D}_2$-$\mathcal{V}_4$
It can be seen that the predicted values generated by TDN-$\mathcal{D}_1$-$\mathcal{V}_4$ nicely track the truth values,
yielding a small error.
Conversely, the predictions of the direction and magnitude of the faults via TDN-$\mathcal{D}_3$-$\mathcal{V}_4$
and TL-FCNN-$\mathcal{D}_2$-$\mathcal{V}_4$ have a significant deviation.
The FE difference among them lies mainly in whether to adopt the decoupling strategy and what activation function in $\mathcal{D}$ to choose.
Compared to the ``Affine'' activation function, the tails of the ``Gaussian'' and ``Square'' exhibit suppression
and amplification effects, i.e., their outputs asymptotically approach 1 and positive infinity,
respectively, as the input increases.
This may lead to the scenario that the predicted fault amplitudes in Fig. \ref{fig_esti} (c) and (d) are smaller and larger than the true values, respectively.
\vspace{-10pt}

\subsection{Three-Tank System Simulation}\label{Sce4:B}
\begin{figure}[!ht]\centering
	\includegraphics[width=5.5cm]{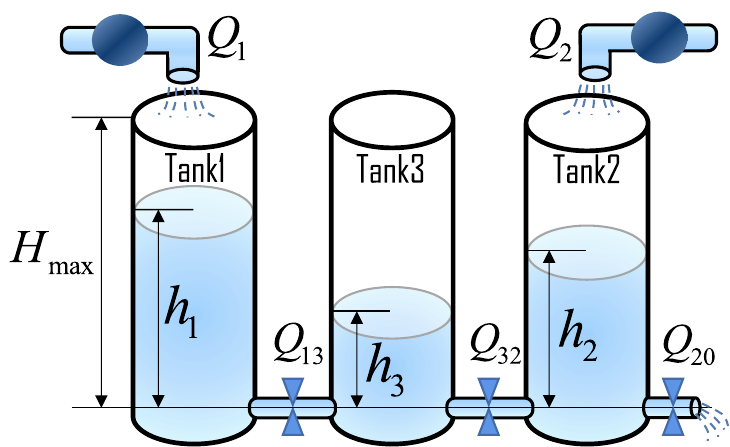}
	\caption{Schematic diagram of TTS simulation.}\label{fig5-2}
\end{figure}

The TTS simulation, depicted in Fig. \ref{fig5-2}, exhibits typical attributes of tanks,
piping, and pumps commonly employed in chemical processes,
and therefore is frequently adopted as a benchmark for laboratory studies.
The TTS simulation utilized in this study is based on the laboratory configuration DTS200,
in conjunction with the parametric depiction by Ding \cite{basin2015finite}.
As shown in Fig. \ref{fig5-2},
there are two pumps feeding material into the tank 1 and 2 with the flow rate of $Q_1$ and $Q_2$, respectively.
Tank 3 is connected to tank 1 and tank 2 through pipes with  cross-sectional mass flow rate $Q_{13}$ and $Q_{23}$.
The state variables $h1,h2,h3$ describes the corresponding water levels of the three tanks. In the case of all states measurable,
the input $u(k)$ and output $y(k)$ of the TTS can be separately expressed by

Its input variables $u(k)$ and measured variables $y(k)$ can be represented separately as
\begin{equation}\label{eq5-7}
    \begin{split}
    &u(k)=[ \begin{array}{cc}
    Q_1(k)   &
    Q_2(k)   
    \end{array}]^T\\
      &y(k)=[\begin{array}{ccc}
     h_1(k)   &
     h_2(k)   &
     h_3(k)
    \end{array}]^T
    \end{split}
\end{equation}

According to Torricelli's law, the nonlinear dynamics of TTS can be formulated by
\allowdisplaybreaks
\begin{equation}\label{eq5-8}
    \begin{split}
    \mathcal{C}{\dot h}_1 &= Q_1 - Q_{13}\\
     \mathcal{C}{\dot h}_2&=  Q_2 +Q_{32}-Q_{20} \\
      \mathcal{C}{\dot h}_3 &= Q_{13}- Q_{32}\\
      Q_{13}&=a_1 \tau {\rm sign}(h_1-h_3)\sqrt{2g\left|h_1-h_3\right|} \\
      Q_{32}&=a_3\tau {\rm sign}(h_3-h_2)\sqrt{2g\left|h_3-h_2\right|}\\
      Q_{20}&=a_2\tau\sqrt{2gh_2},
    \end{split}
\end{equation}
where $\mathcal{C} = 154 ~cm^2$; $\tau  = 0.5~cm^2$ ; $a_1 = 0.46$; $a_2 = 0.6$; $a_3 = 0.45$; $g = 9.8~m/s^2$;
${\rm sign}(\cdot)$ denotes the function to take the sign of the input value.
The detailed meaning of the parameters in \eqref{eq5-8} can be found in reference \cite{dormido2008development}.
\renewcommand{\arraystretch}{1.2}
\begin{table}[!htb]\centering
    \caption{Description of Faults in TTS Simulation}\label{num_faults_TTS}
    \resizebox{\columnwidth }{!}{
    \begin{tabular}{clc}
    \toprule[0.8pt]
    \makecell[c]{Fault ID}  &  \makecell[c]{Description} & \makecell[c]{Location} \\
    \hline\hline \noalign{\smallskip}
    \makecell[c]{Fault01} & \makecell[c]{$f_1(k) =0.005(k-200)+3({1-\lfloor\frac{k}{64}-\frac{5}{4}\lfloor\frac{k}{80}\rfloor\rfloor}) $} & \makecell[c]{$Q_1,Q_2$} \\
    \makecell[c]{Fault02} & \makecell[c]{$f_2(k) = -20$} & \makecell[c]{$Q_1$} \\
    \makecell[c]{Fault03} & \makecell[c]{$f_3(k) = -0.005(k-200)$}& \makecell[c]{$h_1$} \\
    \makecell[c]{Fault04} & \makecell[c]{$f_5(k) = 0.0003(k-200)+0.0006{\rm sin}\frac{k-200}{2\pi}$} & \makecell[c]{$h_2$} \\
    \makecell[c]{Fault05} & \makecell[c]{$f_4(k) = 0.005(k-200)$} & \makecell[c]{$\dot{h_1} $} \\
    \makecell[c]{Fault06} & \makecell[c]{$f_6(k) = -0.0004(k-200)$} & \makecell[c]{$\dot{h_2} $} \\
     \makecell[c]{Fault07} & \makecell[c]{$f_{7}(k) = -0.0004(k-200)$} & \makecell[c]{$\dot{h_3} $}\\
     \makecell[c]{Fault08} & \makecell[c]{$f_{8}(k) = -0.5$} & \makecell[c]{$Q_{13}$}
    \cr
    \hline
    \end{tabular}}
\end{table}
\renewcommand{\arraystretch}{1.2}
\begin{table*}[t]\centering
    \caption{FD Comparison Results of TDNs and Other Four Advanced Methods on the TTS Simulation}\label{TTS_result}
    \begin{tabular}{ccccccccccccccc}
    \toprule
    \multirow{3}{*}{Index} & \multicolumn{7}{c}{$\hat p_{FAR}(\%)$} & \multicolumn{7}{c}{$\hat p_{MDR}(\%)$} \\ 
    \cmidrule(r){2-8} \cmidrule(l){9-15} 
     & \multirow{2}{*}{VAE} & \multirow{2}{*}{K-VAE} & \multirow{2}{*}{AAE} & \multirow{2}{*}{DAE}
     & \multicolumn{3}{c}{TDN} & \multirow{2}{*}{VAE} & \multirow{2}{*}{K-VAE} & \multirow{2}{*}{AAE} & \multirow{2}{*}{DAE} 
     & \multicolumn{3}{c}{TDN} \\ 
     \cmidrule(r){6-8} \cmidrule(l){13-15}
     & & & & &$\mathcal{D}_1$&$\mathcal{D}_2$&$\mathcal{D}_3$& & & & &$\mathcal{D}_1$&$\mathcal{D}_2$&$\mathcal{D}_3$\\
    \hline\hline \noalign{\smallskip}
    $\mathcal{V}_1$ & 0.31 & 0.25 & 0.25 & 0.31 & 0.25 & 0.25 & 0.43 & 4.67 & 7.87 & 3.46 & 3.60 & 3.46 & \textbf{3.45} & 6.30 \\
    $\mathcal{V}_2$ & 0.31 & 0.25 & 0.12 & 0.25 & 0.25 & 0.25 & 0.37 & 4.35 & 4.45 & 3.98 & {3.50} & \textbf{3.46} & \textbf{3.46} & 6.11 \\
    $\mathcal{V}_3$ & 0.37 & 0.31 & 0.31 & 0.25 & 0.25 & 0.25 & 0.37 & 4.29 & 4.43 & 4.32 & \textbf{3.45} & 3.46 & \textbf{3.45} & 6.15 \\
    $\mathcal{V}_4$ & 0.25 & 0.25 & 0.12 & 0.37 & 0.25 & 0.31 & 0.31 & 3.72 & 4.09 & 4.32 & 3.47 & \textbf{3.46} & 3.47 & 5.61 \\
    $\mathcal{V}_5$ & 0.18 & 0.24 & 0.375 & 0.37 & 0.25 & 0.25 & 0.31 & 3.70 & 4.12 & 4.27 & 3.45 & 3.46 & \textbf{3.45} & 5.54 \\
    $\mathcal{V}_6$ & 0.18 & 0.06 & 0.31 & 0.37 & 0.25 & 0.31 & 0.37 & 3.72 & 4.65 & 4.08 & 3.52 & \textbf{3.46} & 3.47 & 5.60\\
    \hline
    \end{tabular}
\end{table*}

To test the FD and FE performance of the proposed method,
eight different faults were introduced to TTS simulation, as described in TABLE \ref{num_faults_TTS}. 
$\left\lfloor \cdot \right\rfloor $ stands for the floor function rounding a number down to the nearest integer.
The first four of them are sensor faults, and the rest four are component faults that simulate tank leakage and pipe blockage.
The variables affected by the component faults can be expressed on the basis of the original one as
\begin{equation}\label{TTS_f}
    \begin{split}
        &\dot{h_i} := \dot{h_i} - a_i \tau f_{i+4} \sqrt{2gh_i}, i = 1,2,3,\\
        &Q_{13} := Q_{13} + a_1 \tau f_8 sign(h_1 - h_3) \sqrt{2g |h_1 - h_3|}.
    \end{split}
\end{equation}

Then, a similar sampling process was performed to obtain the training and test sets in TTS simulation.
A total of 16000 normal samples were collected for offline training and 2000 $\times$ 8 samples (including 200 $\times$ 8 normal samples and 1800 $\times$ 8 faulty samples) for online testing.
In the eight different testing sets,
each containing 200 normal samples followed by 800 faulty samples corresponding to the eight faults defined in TABLE \ref{num_faults_TTS}.
\begin{figure}[t]\centering
    \begin{subfigure}{.5\textwidth}
        \centering
        \includegraphics[width=.88\linewidth]{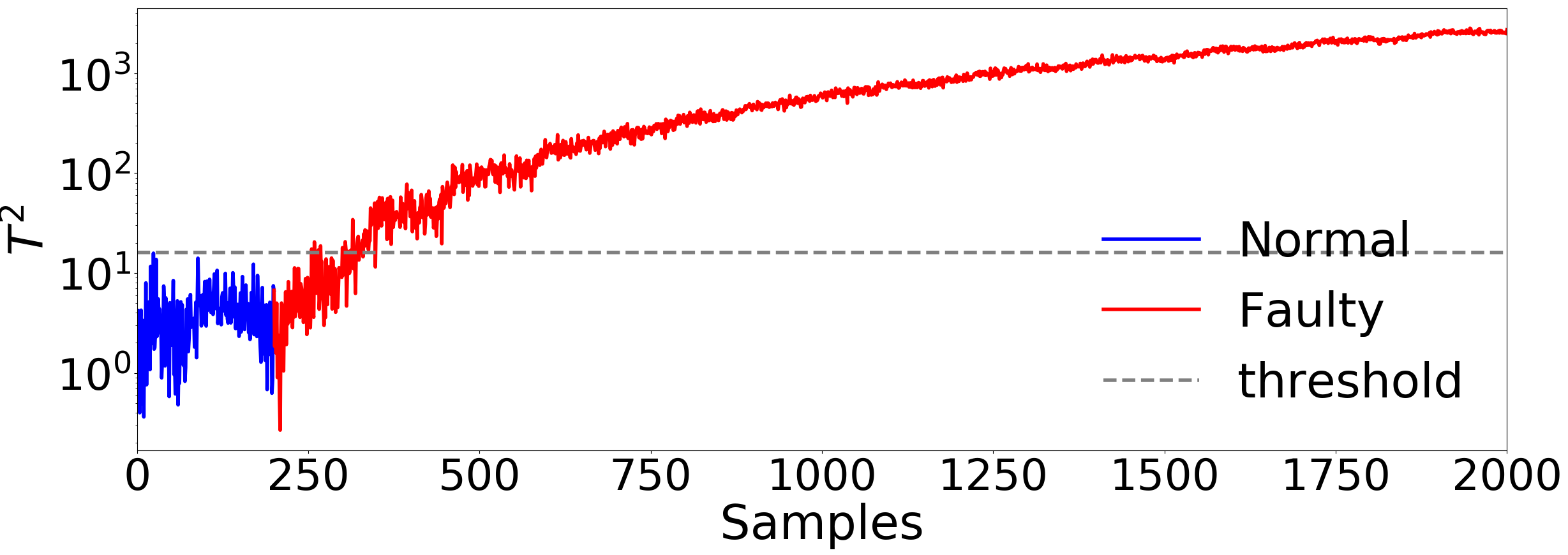}  
        \caption{$Fault\;03$}
        \label{figTTS-a}
    \end{subfigure}
    \newline
    \begin{subfigure}{.5\textwidth}
        \centering
        \includegraphics[width=.88\linewidth]{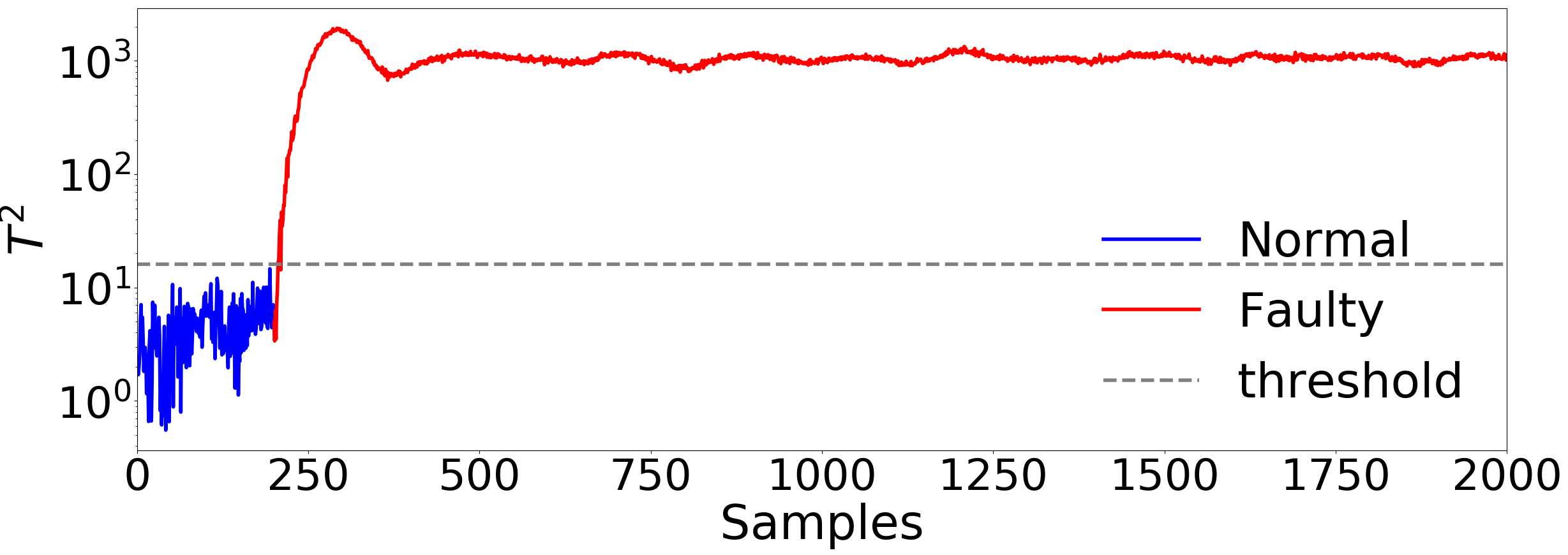}  
        \caption{$Fault\;08$}
        \label{figTTS-b}
    \end{subfigure}
    \caption{Representative FD curves on TTS simulation given by TDN-$\mathcal{D}_2$-$\mathcal{V}_3$}
    \label{figTTS}
    \vspace{-15pt}

\end{figure}

\begin{figure}[t]\centering
	\includegraphics[width=.85\linewidth]{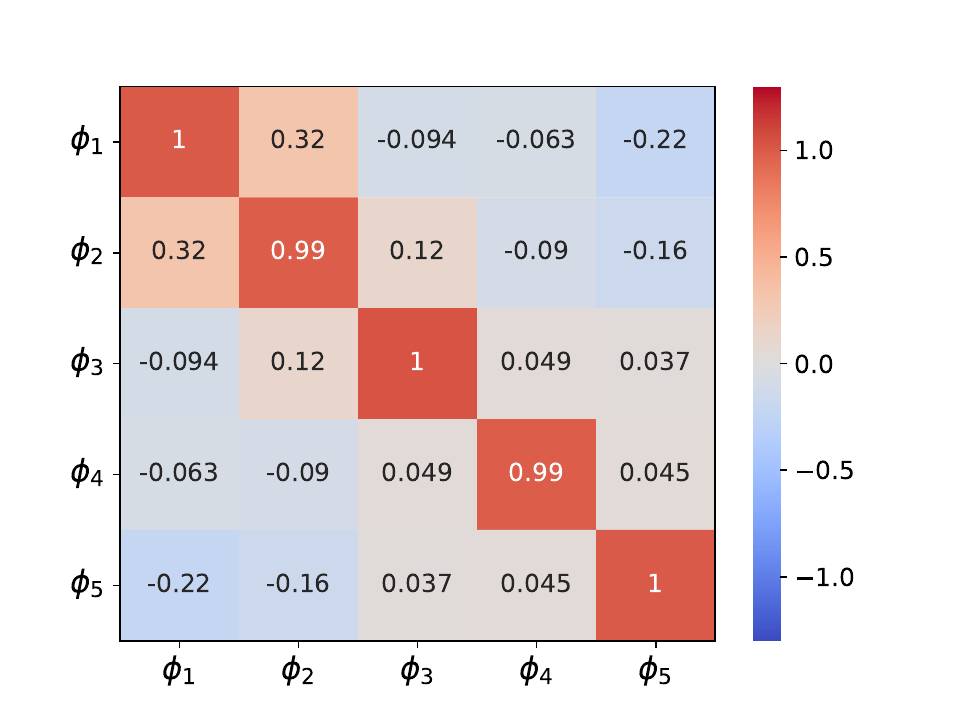}
	\caption{Covariance matrix of TDN-$\mathcal{D}_2$-$\mathcal{V}_3$-generated $\phi(z^n)$}\label{fig_cov_TTS}
    \vspace{-15pt}
\end{figure}

Similarly, the different structures given in TABLE \ref{structures} are also applicable to the TTS simulation by
concatenating the inputs and outputs of the system as $z(k) = [u^T(k)~~y^T(k)]^T$ and feeding them into the network. 
Also, other hyperparameters remain consistent with those in the numerical example.
In this way, 18 TDNs with different network architectures were constructed for FD and FE purposes on TTS simulation.
After ten independent replicate trials,
the mean of AFARs and AMDRs for TDNs and other four advanced models are shown in TABLE \ref{TTS_result}.
Test results and conclusions are similar to those in the numerical example.
TDN-$\mathcal{D}_1$ and TDN-$\mathcal{D}_2$ achieve the best FD performance,
with AFARs no higher than $0.50\%$ and AMDRs no higher than $3.47\%$.
Likewise, the performance of $TDN-D_3$ remains poor,
consistent with that analyzed in the numerical examples.
In addition, DAE demonstrates promising performance in both $\mathcal{V}_2$ and $\mathcal{V}_3$ configurations.
Its overall FD performance is superior to VAE, K-VAE, and AAE, but not as impressive as TDN.

Concretely, Fig. \ref{figTTS} illustrates the FD curves of TDN-$\mathcal{D}_2$-$\mathcal{V}_3$ for 
some representative faults.
It can be seen that the thresholds learned with TDN (dashed lines) are able to correctly classify normal and faulty samples.

Moreover, the covariance matrix of $\phi(z^n)$ generated by TDN-$\mathcal{D}_2$-$\mathcal{V}_3$ is shown in Fig. \ref{fig_cov_TTS}.
One can conclude that it is close to a unitary matrix, thanks to the designed decoupling structure.
To demonstrate the advantages of decoupling, we compare the FE performance of TDN (IDN+VAE) with TL-FCNN (FCNN+VAE) on TTS simulation.

\renewcommand{\arraystretch}{1.2}
\begin{table}[t]\centering
    \caption{Detailed FE Results (ARMSEs) via TDNs}\label{esti_TTS}
    \begin{tabular}{>{\centering\arraybackslash}p{1.6cm}>{\centering\arraybackslash}p{1.6cm}>{\centering\arraybackslash}p{1.6cm}>{\centering\arraybackslash}p{1.7cm}}
    \toprule
    \multirow{2}{*}{\makecell[c]{Structure}} & \multicolumn{3}{c}{TDN-$\mathcal{V}_3$}  \\
    \cmidrule(r){2-4} 
     & $\mathcal{D}_1$ & $\mathcal{D}_2$ & $\mathcal{D}_3$\\
    \hline\hline \noalign{\smallskip}
    $\mathcal{V}_1$ & 2.057 & 2.140 & 13.87 \\
    $\mathcal{V}_2$ & 2.060 & 2.371& 13.77  \\
    $\mathcal{V}_3$ & 2.052 & 2.100& 13.85\\
    $\mathcal{V}_4$ & 2.052 & 2.083 & 13.81 \\
    $\mathcal{V}_5$ & 2.056 & 2.905 & 13.83 \\
    $\mathcal{V}_6$ & 2.055 & 2.440 & 13.81 \\
    \hline
    \end{tabular}
\end{table}
\renewcommand{\arraystretch}{1.2}
\begin{table}[t]\centering
    \caption{Detailed FE Results (ARMSEs) via TL-FCNNs}\label{fcnn_esti_TTS}
    \begin{tabular}{>{\centering\arraybackslash}p{1.6cm}>{\centering\arraybackslash}p{1.6cm}>{\centering\arraybackslash}p{1.6cm}>{\centering\arraybackslash}p{1.7cm}}
    \toprule
    \multirow{2}{*}{\makecell[c]{Structure}} & \multicolumn{3}{c}{TL-FCNN}  \\
    \cmidrule(r){2-4} 
     & $\mathcal{D}_1$ & $\mathcal{D}_2$ & $\mathcal{D}_3$\\
    \hline\hline \noalign{\smallskip}
    $\mathcal{V}_1$ & 5.034 & 50.51 & 15.11 \\
    $\mathcal{V}_2$ & 4.764 & 51.74& 15.02  \\
    $\mathcal{V}_3$ & 4.918 & 51.95& 15.05\\
    $\mathcal{V}_4$ & 5.067 & 67.74 & 14.98 \\
    $\mathcal{V}_5$ & 4.826 & 68.83 & 15.05 \\
    $\mathcal{V}_6$ & 4.991 & 68.40 & 15.00 \\
    \hline
    \end{tabular}
    \vspace{-13pt}
\end{table}

\begin{figure}[!htb]\centering

    \begin{subfigure}{.5\textwidth}
        \centering
        \includegraphics[width=.88\linewidth]{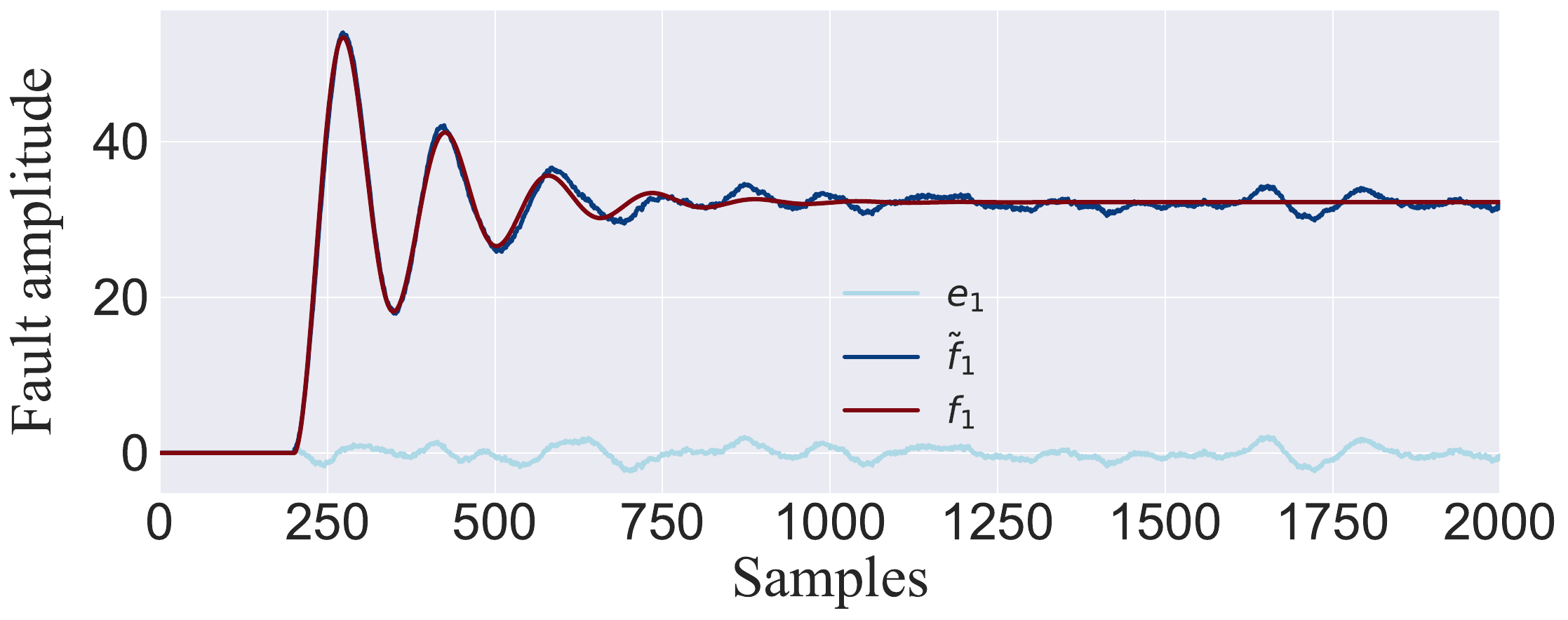}  
        \caption{Estimation of Fault02 using {TDN}-$\mathcal{D}_1$-$\mathcal{V}_3$ (Affine)}
        \label{fig_esti-b}
    \end{subfigure}

    \begin{subfigure}{.5\textwidth}
        \centering
        \includegraphics[width=.88\linewidth]{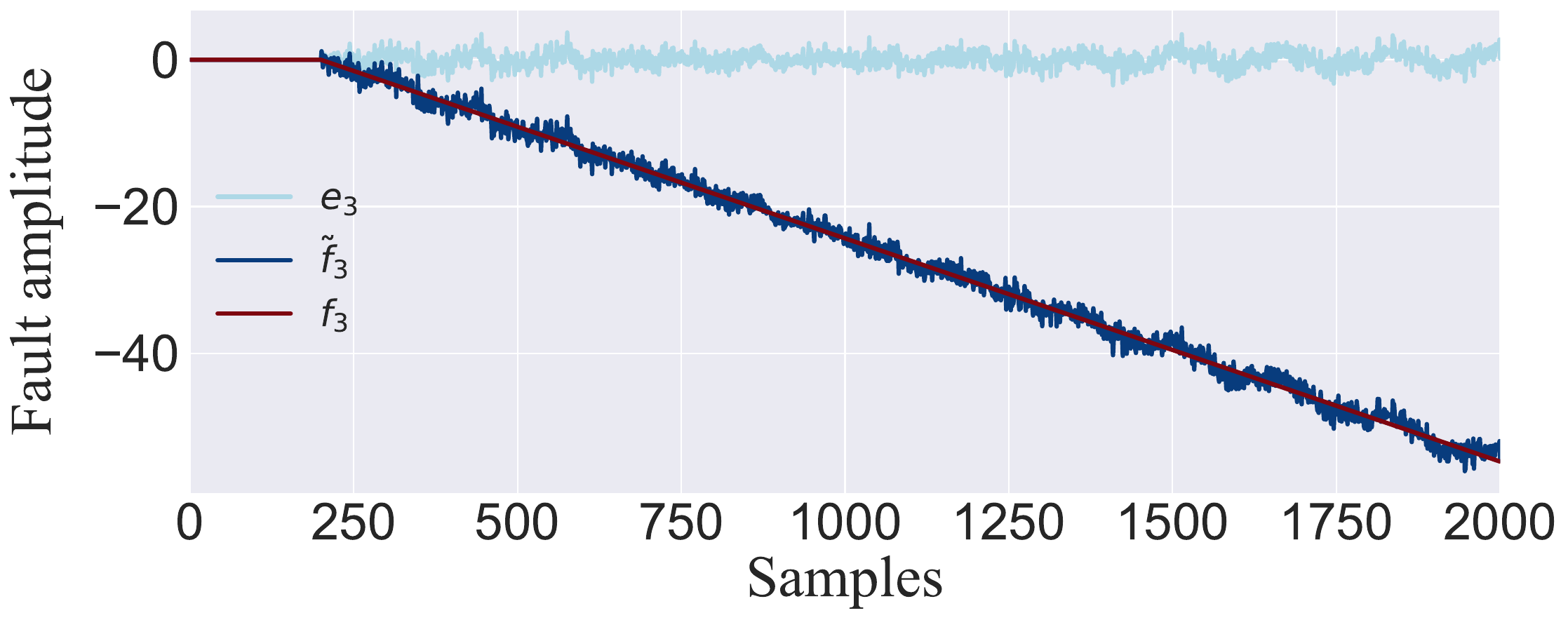}  
        \caption{Estimation of Fault03 using {TDN}-$\mathcal{D}_1$-$\mathcal{V}_3$ (Affine)}
        \label{fig_esti-c}
    \end{subfigure}

    \begin{subfigure}{.5\textwidth}
        \centering
        \includegraphics[width=.88\linewidth]{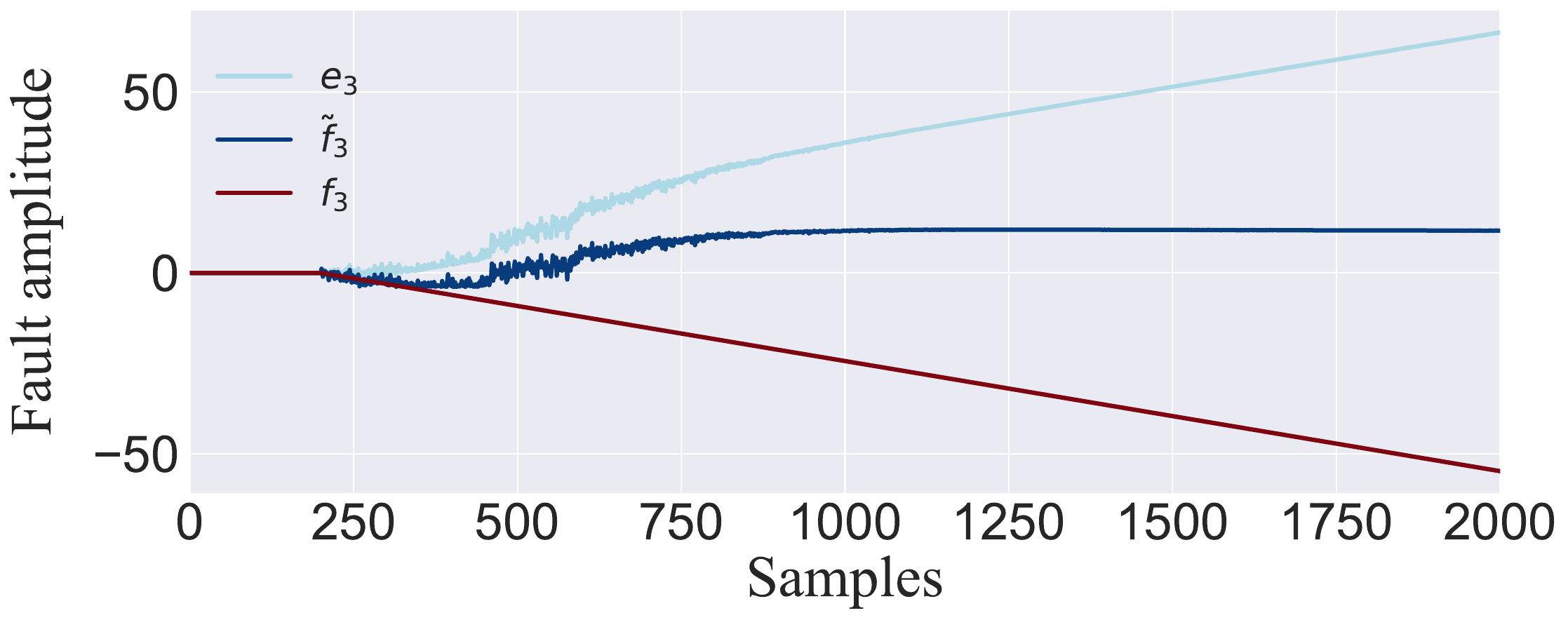}  
        \caption{Estimation of Fault03 using {TDN}-$\mathcal{D}_3$-$\mathcal{V}_3$ (Gaussian)}
        \label{fig_esti-d}
    \end{subfigure}
    \begin{subfigure}{.5\textwidth}
        \centering
        \includegraphics[width=.88\linewidth]{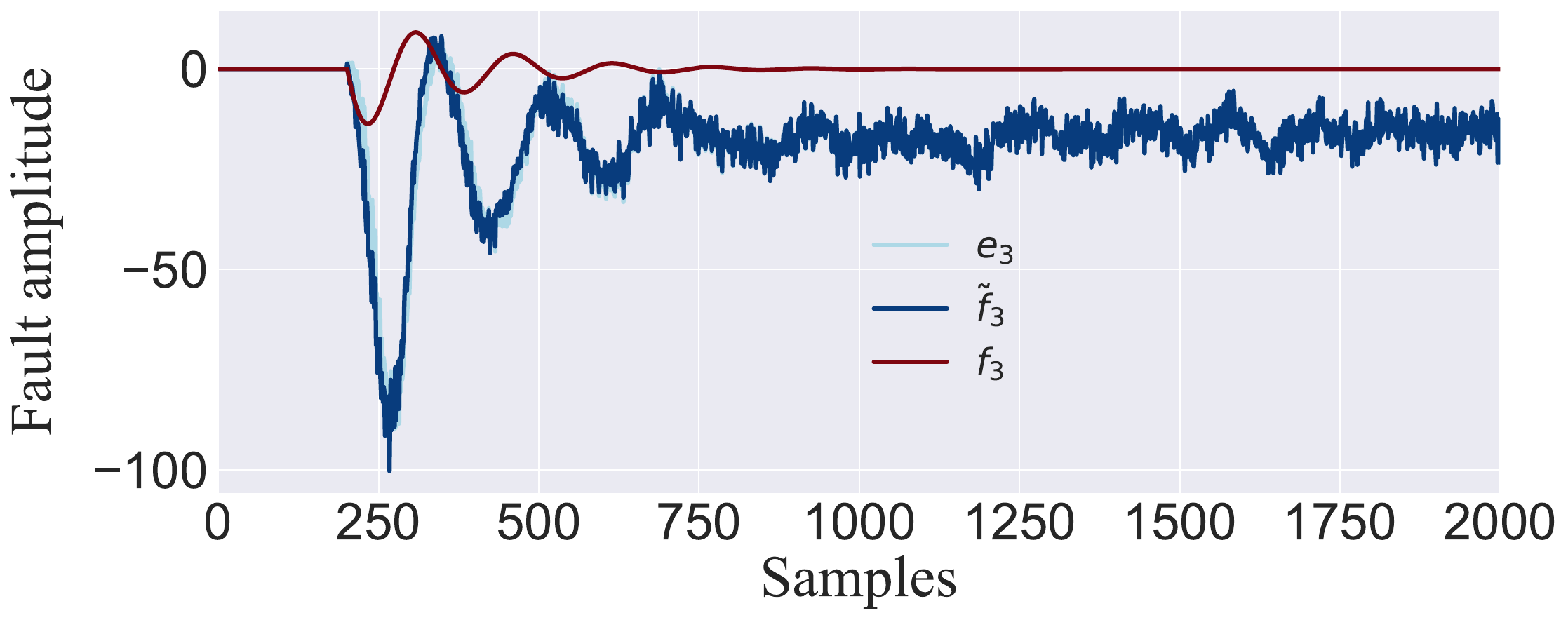}  
        \caption{Estimation of Fault02 using {TL-FCNN}-$\mathcal{D}_2$-$\mathcal{V}_3$ (Square)}
        \label{fig_esti-a}
    \end{subfigure}
    \caption{
        FE curves on TTS simulation given by TDN-$\mathcal{D}_1$-$\mathcal{V}_3$,
    TDN-$\mathcal{D}_3$-$\mathcal{V}_3$ and {TL-FCNN}-$\mathcal{D}_2$-$\mathcal{V}_3$}
    \label{fig_esti_TTS}
    \vspace{-25pt}
\end{figure}

TABLEs \ref{esti_TTS} and \ref{fcnn_esti_TTS} list the ARMSE results for TDN and TL-FCNN under 18 different structures, respectively.
Overall, TDN-$\mathcal{D}_1$ and TDN-$\mathcal{D}_2$ achieved relatively low ARMSEs,
while TDN-$\mathcal{D}_3$, TL-FCNN-$\mathcal{D}_3$ and TL-FCNN-$\mathcal{D}_3$ produced large prediction errors.
Upon comparing the TDNs with different $\mathcal{D}$-structures,
we can observe that the selection of the activation function has a significant effect on the FE results. 
Specifically, the Affine activation function is more suitable for the design of IDNs compared to Gaussian and Square.
This is because the latter two functions inhere flat and steep nonlinear tendencies, respectively,
which can lead to under- or over-prediction of the fault amplitude.
In addition to the activation function, the design of the network structure also has remarkable effects on the FE performance.
By comparing TDN-$\mathcal{D}_1/\mathcal{D}_2$ (with decoupling design) and FL-FCNN-$\mathcal{D}_1/\mathcal{D}_2$
(without decoupling design), it reveals a huge difference existing in their FE results,
especially in the case of the TL-FCNN-$\mathcal{D}_2$, where the results are almost meaningless and unacceptable.

Fig. \ref{fig_esti_TTS} provides detailed fault reconstruction curves by using {TDN}-$\mathcal{D}_1$-$\mathcal{V}_3$ and {TDN}-$\mathcal{D}_3$-$\mathcal{V}_3$,
where light blue, dark blue, and red lines represent the error, estimated values, and true value of the faults, respectively. 
The first two figures present the estimation results by TDN-$\mathcal{D}_1$-$\mathcal{V}_3$,
and the last two figures are offered by {TDN}-$\mathcal{D}_3$-$\mathcal{V}_3$ and TL-FCNN-$\mathcal{D}_2$-$\mathcal{V}_3$, respectively.
Clearly, the fault prediction results of TDN-$\mathcal{D}_1$-$\mathcal{V}_3$ exhibit
a remarkable proximity to the actual fault signal.
In contrast, TDN-$\mathcal{D}_3$-$\mathcal{V}_3$ and
TL-FCNN-$\mathcal{D}_2$-$\mathcal{V}_3$ demonstrate signal compression and amplification, respectively.
Furthermore, both methods predict the fault direction incorrectly, opposite to the expected one.

To summarize, the FD and FE performance of the TDN is verified through a numerical example and a TTS simulation,
which shows more superior capabilities comparing to the other four state-of-the-art methods.
The experimental results indicate the positive contribution of the decoupled structure and its proper activation function selection for better fault diagnosis.

\section{Conclution}\label{Sec5}
In this paper, transfer learning-based input-output decoupled
network is proposed for FD and FE objectives, which consists of an IDN to generate uncorrelated residual variables and a pre-trained VAE to guide the learning of IDN.
In IDN, a clever pre-operation is designed to convert an individual sample into multiple row inputs and then feed into IDN in parallel for structural decoupling.
Then, the proposed IDN is embedded into a TL framework to train its parameters and achieve FE functionality.
During TL phase, normal and simulated faulty samples are co-input into the IDN and subsequently VAE for deriving network outputs.
With the guidance of VAE loss and maximum mean discrepancy loss,
the IDN will learn the state migration mapping from faulty to normal, thereby serving the FE purpose.
Finally, the effectiveness of the proposed method are validated by a numerical example and TTS simulation.
It obtains higher FD and FE performance than other four state-of-the-art methods and an un-decoupled TL-based model.
Furthermore, the results reveal the aggressive contribution of the decoupling in the FE task.
Also, correct activation function selection and network structural design play a crucial role in deep learning-based FD and FE.

\bibliographystyle{IEEEtran}

\bibliography{reference}

\end{document}